\documentclass[preprint,10pt,3p]{elsarticle}

\usepackage{amssymb}
\usepackage{graphicx,multirow,caption,subcaption,multicol,setspace,amsmath,wrapfig}
\usepackage[linesnumbered,ruled]{algorithm2e}
\usepackage{algorithmic}
\usepackage{adjustbox}
\usepackage{booktabs,caption}
\usepackage[flushleft]{threeparttable}
\usepackage{lineno}
\usepackage{enumitem}
\setlist{nosep}
\usepackage{rotating,caption}
\usepackage[table]{xcolor}
\usepackage{soul,xcolor}
\usepackage{amsfonts}
\graphicspath{ {Images/} }

\journal{Pattern Recognition}

\begin{document}
\begin{frontmatter}

\title{A Two-Dimensional (2-D) Learning Framework for Particle Swarm based Feature Selection}

\author[label5]{Faizal Hafiz\corref{cor1}}
\address[label5]{Department of Electrical \& Computer Engineering, The University of Auckland, Auckland, New Zealand}
\ead{faizalhafiz@ieee.org}
\cortext[cor1]{Corresponding author}

\author[label5]{Akshya Swain}

\author[label5]{Nitish Patel}

\author[label1]{Chirag Naik}
\address[label1]{Sarvajanik College of Engineering \& Technology, Surat, India}

\begin{abstract}

This paper proposes a new generalized two dimensional learning approach for particle swarm based feature selection. The core idea of the proposed approach is to include the information about the subset cardinality into the learning framework by extending the dimension of the velocity. The 2D-learning framework retains all the key features of the original PSO, despite the extra learning dimension. Most of the popular variants of PSO can easily be adapted into this 2D learning framework for feature selection problems. The efficacy of the proposed learning approach has been evaluated considering several benchmark data and two induction algorithms: \textit{Naive-Bayes} and \textit{k-Nearest Neighbor}. The results of the comparative investigation including the time-complexity analysis with GA, ACO and five other PSO variants illustrate that the proposed 2D learning approach gives feature subset with relatively smaller cardinality and better classification performance with
shorter run times.

\end{abstract}


\end{frontmatter}

\section{Introduction}
\label{sec1:intro}

In recent years, the feature selection problem has been a major concern amongst the researchers in the machine learning field due to availability of huge volume of data from diverse fields such as finance, bio-medical, physical sciences and consumer electronics, etc. The feature selection problem arises from the fundamental question of the machine learning: \emph{How many input features are required to sufficiently capture the characteristics of a data pattern/model?} In the absence of this information large number of input features are used to represent a pattern which often leads to inclusion of many redundant, irrelevant or noisy features. The determination of effective feature subset has been a fundamental problem in machine learning and a topic of active research since last few decades \cite{Marill:Green:1963}.

Recognition of a pattern by a machine learning method involves induction of hypotheses by an induction algorithm which maps the input features to the output class/label. It is highly desirable to induce a classifier using fewer input features as possible for various reasons. For example, in most cases, only limited samples are available for training, whereas training samples required to achieve required accuracy rise exponentially with the number of input features \cite{Blum:Langley:1997}. Further, earlier studies \cite{Blum:Langley:1997,Blumer:Ehrenfeucht:1987} suggest that good generalization capabilities can be achieved by a smaller feature subset following the \emph{principle of parsimony} or \emph{Occam's razor}. Moreover, there may be constraints imposed on number of input features due to limited resources, such as number and/or cost of measurements associated with features, storage requirements. 

To understand the feature selection problem, consider a dataset with `$n$' input features ($U=\{u_1 \dots u_n \}$) and `$m$' output labels ($V = \{ v_1 \dots v_m \}$). The task of the induction algorithm is to induce hypotheses in the classifier using the learning data pairs $\{U,V \}$ which can later be used to determine appropriate label, $v_k \in V$, corresponding to any future input pattern. The objective is to find a subset of input features,`X' ($X \subset U$, $|X|<n$), through which this task can be accomplished with same or improved accuracy. This can further be represented as follows,
\begin{linenomath*}
\begin{align}
    \label{eq:fsp}
    J(X) = \max \limits_{Y \subset U, \ |Y|<n} J(Y) 
\end{align}
\end{linenomath*}
where, `$J(\cdotp)$' is a criterion function which estimates the `\emph{goodness}' of the given subset. An exhaustive search of all possible feature subsets to solve (\ref{eq:fsp}) requires the examination of large number of feature subsets ($\sum \limits_{k=1}^{n-1} \binom n k \approx 2^n$), which often becomes intractable even for moderate size problems. The feature selection problem is NP-hard \cite{Blum:Langley:1997,Cover:Van:1977,Guyon:Isabelle:2003} and its optimal solution is not guaranteed unless all possible $2^n$ feature subsets are examined \cite{Cover:Van:1977}. The optimal solution of this problem requires the selection of both the subset size (\emph{cardinality})  and the features themselves.

Most of the methods which have been proposed to address the feature selection problem  can be broadly be classified into two categories: \emph{deterministic} and \emph{meta-heuristic}. Majority of deterministic search methods such as \emph{sequential search} \cite{Marill:Green:1963,Whitney:1971,Somol:Pudil:1999,Reif:Shafait:2014}, \emph{branch and bound} \cite{Narendra:1977,Nakariyakul:2014} require \emph{monotonic} criterion function, $J(\cdotp)$, and/or neglects correlation among features. Therefore, some of the recent search methods uses alternate search approach based on meta-heuristics \cite{Xue:Zhang:2016}. The meta-heuristic search methods have proven to be very effective on various discrete and combinatorial problems. If properly adapted, they can provide optimal solution to the feature selection problem. The pioneer effort in this direction was the application of Genetic Algorithm (GA) for the feature selection problem in \cite{Siedlecki:Sklansky:1989}. Apart from GA, various other meta-heuristics search such as Tabu Search (TS), Ant Colony Optimization (ACO) and Particle Swarm Optimization (PSO) have been applied to the feature selection task \cite{Xue:Zhang:2016,De:Fontanella:2014,Banka:Dara:2015,Chen:Chen:2013,Yu:Gu:2009}. Note that all of the meta-heuristic search paradigms have to be adapted for the feature selection task. Unlike PSO, the other search paradigms like GA, ACO can be adapted comparatively easily without any major changes in their learning strategies. For example, GA can be applied with a binary string representation and graph representation can be used for ACO. On the other hand, the similar task is quite challenging with canonical PSO due to \emph{euclidean distance} based learning at its core\cite{Kennedy:Eberhart:1995}. Nevertheless, in the recent research, PSO is preferred due to its simplicity, its ability to avoid local minima and it does not require any heuristic information other than the criterion function, $J(\cdotp)$ \cite{Xue:Zhang:2016}.

Unlike other evolutionary search paradigms, PSO has a single learning mechanism; known as \emph{velocity update}. In canonical PSO, the velocity update involves evaluation of the \textit{euclidean distance} of the particle from its learning exemplars, referred to as \emph{learning} \cite{Kennedy:Eberhart:1995,Hafiz:Abdennour:2013}. The next move of the particle on the search landscape is evaluated based on this learning. Since the euclidean distance does not convey useful information in the discrete domain, a binary version of PSO (BPSO) was proposed in \cite{Kennedy:Eberhart:1997}, where the velocity is represented as the \emph{selection probability}. However, BPSO has several limitations which severely affect its search performance (discussed at length in Subsection \ref{sec:PSOreview}). Moreover, BPSO is intended to be a general search paradigm for any discrete problem and for this reason does not contain any specific search mechanism to cope with the feature selection problem. Most of the recent research on PSO based feature selection are either application of BPSO or some extension of it. The new search strategies dedicated to feature selection problem is still an open issue \cite{Xue:Zhang:2016}.

The objective of this study is to bridge this gap by proposing a new learning framework for the PSO and its variants. The proposed learning framework is designed as \emph{generic learning framework} which can be used to adapt any PSO variant for the feature selection problems. Since its introduction, PSO has attracted many researchers and over the years many PSO variants have been proposed to improve the performance of the original algorithm albeit the research has been mostly restricted to the `\textit{continuous domain}', \textit{i.e.}, $x \in \mathbb{R}$. The introduction of a generalized learning framework can help in transferring most of the existing PSO related research from \textit{continuous domain} ($x \in \mathbb{R}$) to the feature selection problem ($x \in \mathbb{N}$). This is the main motivation of this research. 

The search for the optimal feature subset involves two aspects; selection of both \emph{cardinality} and \emph{features}. However, most of the search methods focus only on the significance of feature/feature subset and to the best of our knowledge, none of the search methods directly exploit the information on subset cardinality to guide the search process. In the proposed learning framework, information about both the cardinality and features are jointly exploited to effectively guide the search. Contrary to the earlier practice of storing only selection probabilities of \emph{features} in $n$-dimensional \emph{velocity} vector, in this work, the velocity records the \emph{selection likelihood} of both \emph{cardinality} and \emph{features} in a two-dimensional matrix. Due to this distinctive quality, the proposed learning framework is named `\emph{2D learning approach}'. Moreover, a simple method is proposed to update the cardinality and feature selection likelihoods and generate a new feature subset based on this comprehensive information.

The efficacy of the proposed approach is evaluated on wide variety of real-life datasets obtained from UCI Machine Learning repository \cite{UCI:2013}. Note that the search landscape of the feature selection problem is jointly defined by the dataset and the induction algorithm used to induce the classifier. For this reason, two widely used induction hypotheses based on \emph{Naive-Bayes} and \emph{k-Nearest Neighbor} (k-NN) \cite{Witten:Frank:Hall:2016} have been used to induce the classifier. Further, the well known PSO variant, \emph{Unified Particle Swarm Optimization} (UPSO) \cite{UPSO1}, is adapted for the feature selection problem using the 2D learning approach. The performance of the adapted UPSO (2D-UPSO) is compared with GA and five other PSO based feature selection methods. 

The rest of the manuscript is organized as follows: the brief overview of the feature selection methods is provided in Section \ref{sec:review}, followed by the detailed description of the proposed 2-D learning approach in Section \ref{sec:2-D}. The application of 2-D learning approach to adapt PSO variants is illustrated in Section \ref{sec:adapt}. The experimental setup, compared algorithms and results are discussed in Section \ref{sec:setup} and \ref{sec:res}, respectively. Finally, the conclusions of this study are discussed in the Section \ref{sec:conclusion}.

\section{Brief Review of Feature Selection Methods}
\label{sec:review}

Since, in this study a new feature selection method has been proposed in a 2D learning framework, it is appropriate to briefly discuss the search behavior of the existing feature selection methods for the sake of completeness. Most of the existing feature selection methods can be classified into two distinctive categories on the basis of the nature of search: \emph{deterministic} vs. \emph{meta-heuristic}.  Another important distinction arises from the different approaches to evaluate the criterion function, $J(\cdotp)$ (\emph{filters} vs. \emph{wrappers}).

\subsection{Deterministic vs. Meta-Heuristic Search}
Over the years several deterministic search methods have been proposed such as sequential search \cite{Marill:Green:1963,Whitney:1971,Somol:Pudil:1999,Reif:Shafait:2014}, \emph{branch and bound} (BAB) \cite{Narendra:1977,Nakariyakul:2014}. These methods are \emph{deterministic}, since for a given dataset, each independent run of these methods will provide same solution. The core idea of the sequential search is to operate on a single feature, \textit{e.g.}, forward search (SFS) \cite{Whitney:1971} starts with an empty set and a single feature is included in each step where as in backward search (SBS) \cite{Marill:Green:1963}, the search begins with all the features and a single feature is discarded in each step. The drawback of this approach is the \emph{nesting effect}, \textit{i.e.}, once the feature is included/excluded it cannot be discarded/included. To overcome this problem, `\emph{plus-$l$-minus-$r$}' and \emph{floating} search \cite{Somol:Pudil:1999,Reif:Shafait:2014} methods were suggested. The common drawback of the sequential search is the emphasis on an isolated feature which completely ignores the correlation among the features. Due to correlation, an isolated insignificant feature may become very effective when considered with others \cite{Toussaint:1971}. Further, most of deterministic search methods operate on strict assumption of \emph{monotonic} criterion function, \textit{i.e.}, adding a feature will always lead to improvement.  This assumption is impractical as in many cases, due to limited training samples larger input feature subset often leads to over-fitting and deteriorates the classifier's performance. Moreover, \textit{a priori} selection of subset size ($d$) is required for most of the deterministic methods, consequently the search space reduces to only $\binom n d$ subsets instead of all possible $2^n$ subsets. Hence, it is highly possible that the optimal feature subset is not even included in the search space. 

The meta-heuristic search methods such as Genetic Algorithm (GA), Tabu Search (TS), Particle Swarm Optimization (PSO), Ant Colony Optimization (ACO) can overcome most of the limitations exhibited by deterministic methods. These search methods are \emph{stochastic} in nature and hence their solutions are non-deterministic, \textit{i.e.}, due to their inherent \textit{randomness}, different runs may provide different solution for the same dataset. However, the chances of avoiding local minima are higher with the meta-heuristic search due to their inherent stochastic nature. Moreover, even-though not guaranteed, the meta-heuristic search can provide optimal solution, where as the deterministic methods will always provide sub-optimal solution. In addition, most of the meta-heuristic methods such as GA, ACO and PSO are in essence population based search. The immediate advantage of this is \emph{implicit parallelism}, \textit{i.e.}, each population member can explore subset of different cardinality. Hence, \textit{a priori} selection of subset size is not necessary.  Consequently, during search process, the entire solution space containing $2^n$ subsets is sampled. Therefore, the chances of reaching to global optimal solution are comparatively higher. Further, in most of the meta-heuristic methods, the search progresses on the basis of \emph{significance/relevance} of the entire subset, instead of a single feature. Thus if properly designed, the issue of \emph{feature correlation} can be effectively addressed. 

\subsection{Filter vs. Wrapper}

The other important characteristic which distinguishes the  search methods is the approach which is used to evaluate the \emph{goodness} of a feature/feature subset or \emph{criterion function}, $J(\cdotp)$. The \emph{filter} approach relies on a metric derived from \emph{statistical}, \emph{information theory} or \emph{rough sets} to estimate $J(\cdotp)$ \cite{Xue:Zhang:2016,Wu:Yu:2013,Naghibi:Hoffman:2015,Zeng:Zhang:2015,Jiang:Sui:2015,Freeman:Kulic:2015,Senawi:Billings:2017}. Since, $J(\cdotp)$ is estimated indirectly, the induction of the classifier is not required. Hence, the \emph{filters} are computationally more efficient. Moreover, the subsets derived from \emph{filters} are \emph{generalized}, \textit{i.e.}, the subsets are independent of the induction algorithm and can be used with any induction algorithm. However, the \textit{filters} are limited by the assumption of the ideal Bayesian classifier. Each induction algorithm has its own specific traits which significantly affects the performance of the induced classifier and optimal feature subset. This is ignored in the filter approach. The \emph{search landscape} of the feature selection problem is conjointly defined by the dataset and the induction algorithm. This is the main motivation behind \textit{wrapper} approach. In the wrapper approach, a classifier is induced for each subset under consideration and resulting accuracy is used as the criterion function, $J(\cdotp)$. The wrapper approach is more precise but computationally involved. 

The selection of the either approach involves trade-off in either \textit{precision} or \textit{complexity}. The search dimension is the major factor that influences this decision. As a compromise, for very large datasets, hybrid of filter and wrapper is often used in which filters are applied first to reduce the dimensionality and wrappers are applied on the reduced subset \cite{Guyon:Isabelle:2003}. In this work, we have experimented with the medium size datasets ($ n \leq 1000$). Hence, without the loss the generality the meta-heuristics are applied as \textit{wrappers}. Note that the proposed approach can easily be extended to the \textit{filters} with few modification.

\subsection{Feature Selection through PSO}
\label{sec:PSOreview}
 
The aim of this section is to provide brief review of various learning approaches employed in PSO in the context of the feature selection problem. The detailed review of PSO based feature selection methods has been discussed in \cite{Xue:Zhang:2016}.

The learning approach employed in most of the PSO based feature selection methods can be classified into three categories: (1) \textit{euclidean distance based} \cite{Kennedy:Eberhart:1995} (2) \textit{probabilistic} \cite{Kennedy:Eberhart:1997} and (3) \textit{Boolean logic} based \cite{Gunasundari:2016}. In the canonical PSO \cite{Kennedy:Eberhart:1995}, \textit{euclidean distance} from the learning exemplar is used as a \textit{learning}, as illustrated by the following velocity update rule for the $d^{th}$ dimension of the $i^{th}$ particle,
\begin{linenomath*}
\begin{align}
    \label{eq:stdpso}
    v_{i,d} & =(\omega \times v_{i,d}) + c_1r_1 \times (pbest_{i,d} - x_{i,d}) + c_2r_2 \times (gbest_d-x_{i,d}) \\
    \label{eq:stdpso:pos}
    x_{i,d} & = x_{i,d} + v_{i,d}
\end{align}
\end{linenomath*}
where, $\omega$ is inertia weight, $c_1, c_2$ are acceleration constants and $r_1,r_2$ are uniform random numbers which lie in [0,1]. The euclidean distance based learning has been used for the PSO based wrappers in \cite{Lin:Ying:2008,Xue:Zhang:2013,Xue:Zhang:2014}; where a \textit{threshold} was used to decide whether or not to include the feature. However, this approach is not recommended for the feature selection problem as the euclidean distance does not convey meaningful information in the discrete domain. The Boolean logic based learning relies on bit wise logical operations to learn from the exemplar. The major drawback of this approach is the need to tune the velocity limit, $v_{max}$, defined in terms of cardinality of the particle. Further, this approach is limited by the scalability problem \cite{Gunasundari:2016}.

Perhaps for this reason, majority of PSO based feature selection methods \cite{Chuang:Chang:2008,Unler:Murat:2010,Bae:Yeh:2010,Liu:Wang:2011,Ghamisi:2015a,Ghamisi:2015b, Akkasi:2016,Zhang:Gong:2017} are based on the \textit{probabilistic learning}. To handle the discrete problems, \textit{Kennedy and Eberhart} proposed the probabilistic learning for the binary PSO (BPSO) in which the velocity $v_{i,d}$ serves as \textit{selection likelihood} of the $d^{th}$ dimension. In the BPSO, particles are represented as binary strings and the velocity update rule is similar to (\ref{eq:stdpso}). However, instead of position update rule (\ref{eq:stdpso:pos}), a transformation function, `$s(\cdotp)$', is used to determine the probability of inclusion of the $d^{th}$ bit ($x_{i,d}$) from its velocity ($v_{i,d}$),
\begin{linenomath*}
\begin{align}
    \label{eq:s}
    s(v_{i,d}) & =\frac{1}{1+e^{-v_{i,d}}}\\
    x_{i,d} & = \begin{cases}
                1, \ \ if \ s(v_{i,d})>r \\
                0, \ otherwise
              \end{cases}, \text{`$r$' is a uniform random number}, r \in [0,1] \nonumber
\end{align}
\end{linenomath*}
Despite its simplicity, BPSO has several limitations which can adversely affect the search process. Especially, due to non-monotonic nature of the transformation function $s(\cdotp)$ in (\ref{eq:s}), even a major increase in $v_{i,d}$ does not translate into significant improvement in the probability after particular threshold ($|v_{i,d}| \geq 2.5$). For example, consider the velocity of two dimensions/features $j$ and $k$ of the $i^{th}$ particle, $v_{i,j}=3$ and $v_{i,k}=5$. According to (\ref{eq:s}), their corresponding probabilities will be $0.95$ and $0.99$, respectively. This may lead to inclusion of many redundant features. In addition, the search behavior of BPSO is very sensitive to velocity limit, $v_{max}$ \cite{Kennedy:Eberhart:1997}. The proper selection of $v_{max}$ is essential to balance the \textit{exploration} and \textit{exploitation} of the search space. Further, unlike the canonical PSO, the determination of the new position is completely independent of its current position. 

Note that, even-though BPSO employs a new interpretation of velocity and new mechanism for the position update, the velocity update rule is unaltered, \textit{i.e.}, the euclidean distance is still being used to derive the learning. For this reason, the selection likelihood of the common features (features that are included in both learning exemplar and particle) are likely to decrease during the velocity update. As seen in (\ref{eq:stdpso}), the selection likelihood of the features, which are present in the exemplar ($pbest$ or $gbest$) but not in the particle position ($x$), will increase. However, the likelihood of the common features will decrease. This is not desirable, as the features included in the learning exemplars represent a subset with better performance. 

\section{Proposed Learning Approach}
\label{sec:2-D}

{ The core of idea of PSO is the search directed on the basis of information gained through `\textit{self-learning}' and `\textit{social-learning}' from peers. Usually, the `\textit{learning}' in most of the existing PSO based approaches is focused on identifying the promising features. However, as mentioned earlier, the selection of feature subset entails two decisions: 1) \textit{Which features to select?} and 2) \textit{How many features to select?} Note that the existing learning approaches only address the first decision and therefore are not effective in removing the redundant/irrelevant features. Therefore, the obvious question which arises is whether it is possible to include the information about the number of features. The philosophy of the proposed approach is to answer this question by including the information about cardinality into the learning process.

By extending the learning dimension to include the information about the cardinality, both decisions can be taken efficiently to form an effective feature subset. Through this additional learning dimension, the particle can learn both about the features and cardinality from the learning exemplars. In each iteration, the particle can effectively reduce search space by learning about the beneficial cardinality, which is expected to effectively discard the redundant features and improve the search performance. Further, since the particles are represented as binary vectors, the cardinality information can easily be obtained with less computations.

This is achieved by embedding the following key elements into the proposed new 2D learning framework,} 
\begin{itemize}
    \item A simple method is developed to extract the \textit{learning} from the exemplar to \textit{replace} the Euclidean distance based learning.
    \item The latent information on \textit{subset cardinality} is exploited and stored by extending the dimension of the velocity.
    \item A new method is developed to derive new position of the particle through inclusive learning on cardinality and feature. 
    \item All the key characteristics of canonical PSO are retained, despite the additional learning dimension. As a result, the proposed learning framework is \textit{generic} and it can be used to adapt any PSO variant for feature selection problems.
\end{itemize}

The details about the proposed learning approach is described in the following: 

The learning in PSO is achieved through velocity update wherein a particle gains information from its learning exemplars and its own memory. As discussed in Section \ref{sec:PSOreview}, in the most PSO variants the feature selection problem is addressed by storing probabilities of the features in an $n$-dimensional velocity vector, whereas the information about subset cardinality, which is implicitly present, is not considered. The proposed learning approach explicitly utilizes this information by representing it in an another dimension in the velocity matrix. In the proposed learning approach, a novel two dimensional ($2 \times n$) velocity matrix is used (instead of an n-dimensional vector) to represent both the selection likelihood of \emph{subset cardinality} and \emph{individual feature}. In each iteration, the learning sets are derived from the learning exemplars to update these selection likelihoods. 

In the present study, for a dataset having $n$ number of features, each particle is represented as an `$n$'-dimensional binary string. The presence of `$1$' in the string indicates that the feature is selected and `$0$' indicates otherwise. The number of `$1$' in the binary string denotes the subset cardinality.

As an example, consider the  $i^{th}$ particle ($x_i$) for a dataset having five number of features ($n=5$) with personal ($pbest_i$) and group best ($gbest$). Let,
\begin{linenomath*}
\begin{align*}
    x_i=\{1 \, 0 \, 1 \, 0 \, 1\}, \; \; \; pbest_i= \{0 \, 1 \, 0 \, 0 \, 1\} \;\;\; \text{and} \;\;\; gbest = \{1 \, 1 \, 0 \, 1 \, 0\}
\end{align*}
\end{linenomath*}
Note that in $x_i$, the `$1$' are located at positions 1,3 and 5. This implies that out of five features, features $\{1, 3, 5\}$ are selected in the $i^{th}$ particle. The same can be extended for $pbest_i$ and $gbest$. Further, the binary string representing the particle also contains the information about the cardinality of the feature subset (denoted as $\xi$) which is being exploited in the present study. For the particle considered in this example, the total number of `$1$' equals to three and hence the cardinality of  $x_i$, denoted as ($\xi_i$) equals 3. Most of the PSO based approach \cite{Kennedy:Eberhart:1997,Gunasundari:2016,Lin:Ying:2008,Xue:Zhang:2013,Xue:Zhang:2014,Chuang:Chang:2008,Unler:Murat:2010,Bae:Yeh:2010,Liu:Wang:2011,Ghamisi:2015a,Ghamisi:2015b,Wang:Yang:2007,Yeh:Chang:2009,Unler:Murat:2011} do not utilize this information about the subset cardinality. Given that the size of the optimal subset is not known \emph{a priori}, inclusion of the subset cardinality in the learning can significantly improve the search performance of the swarm. For this reason, in the proposed learning approach, the velocity stores the probability of both subset cardinality ($\xi$) and features in a two dimensional matrix of size $(2 \times n)$ as follows:  
\begin{equation}
\label{eq:vel}
    v=\begin{bmatrix} p_{11} & \dots & p_{1n} \\ 
                        p_{21} & \dots & p_{2n} \end{bmatrix}
\end{equation}
The elements in the first row of the velocity matrix ($p_{11},\dots,p_{1n}$) represent the selection probability of the subset cardinality, i.e., $p_{1,j}$ represent the likelihood that a subset of size (cardinality) `$j$' will be selected from $n$-number of features. For example, for a dataset with five features, if $p_{13}=1.35$, it implies that the likelihood of selecting three number of features out of five is $1.35$.
 
The elements in the second row of the velocity matrix, i.e., `$p_{21},\dots,p_{2n}$' represents the selection likelihoods of individual features, i.e., $p_{2,j}$ represents the likelihood of selecting $j^{th}$ feature. For example, if $p_{23}=0.57$, it implies that the likelihood of selecting the third feature equals to $0.57$. 

For each of the particle, new position is determined based on the selection likelihoods stored in a similar two-dimensional velocity matrix. The position update based on the proposed approach is discussed at length in Subsection \ref{sec:position}.  

After representing the velocity and position following the above procedure, the next step is to develop a new velocity update mechanism. The major challenge in this process is to update
the selection likelihood of the cardinality and features in each iteration according to cognitive and social learning exemplars. In the Canonical version of the PSO, Euclidean distance from the exemplar is often used as a learning concept as illustrated in (\ref{eq:stdpso}). However, the same learning concept cannot be extended to the problems in  discrete domain as the Euclidean distance does not convey useful information when the particle is represented as a binary string. It is therefore essential to develop a new learning approach that can extract learning from exemplar about both subset cardinality and features. This is the main motivation for the proposed learning approach.

In the present study, learning is performed in two stages; in the first stage of learning, the information about subset cardinality is acquired for the cardinality learning set, `$\varphi$'. During the second stage of learning, the selection likelihood of the features is determined and stored in the learning set, `$\psi$'. Once the learning is complete, both learning sets ($\varphi$ and $\psi$) are combined into a two dimensional matrix of size ($2\times n$) to derive the final learning set, $L$. The first row of $L$, corresponds to the subset cardinality and the second row corresponds to the individual features. In each iteration,the velocity ($v$) of each particle is updated through learning sets `$L$' derived from the learning exemplars.

A separate learning set is required for each exemplar such as $pbest$ or $gbest$. In this work, learning set derived from $pbest$ and $gbest$ is denoted with subscript `${cog}$' and `${soc}$', respectively. Since in the Canonical PSO, the current position of the particle contributes to the new position, as illustrated by (\ref{eq:stdpso}), in the proposed learning approach this property is retained by introducing  another learning set, `$L_{self}$'. This learning set is derived from the current position of the particle ($x$). The following subsections provide more details on proposed learning stages and velocity update process.

\subsection{Learning for Subset Cardinality}

The objective of this stage is to learn about the subset cardinality from the learning exemplars such as $pbest$ and $gbest$. For this purpose, consider a binary string `$\alpha$' of length $n$, which could represent either the position $x$ of the particle, or the learning exemplars such as $pbest$ or $gbest$. 

The first step is to evaluate the cardinality of the subset represented by a binary string. The subset cardinality is obtained by counting the number of `$1$' in a binary string, since the bits with `$1$' in the binary string indicates that the corresponding feature is selected. The number of `$1$' in a given binary string $\alpha$, denoted as $\xi_\alpha$, can be evaluated using \emph{`zero-norm'} or \emph{Hamming distance} from the null vector.

The next step is to generate an $n$-dimensional learning set, $\varphi_\alpha$, corresponding to the cardinality, $\xi_\alpha$. Note that all the bits of $\varphi_\alpha$ are zero except $\xi_\alpha^{th}$ bit (from the most the significant bit) which is set to `$1$'. For example, for a binary string $\alpha=\{ 1 \ 0 \ 1 \ 1 \ 0 \}$, the cardinality, $\xi_\alpha=3$ and therefore the third bit of the string $\varphi_\alpha$ (which represents cardinality learning set of $\alpha$) is set to 1, \textit{i.e.}, $\varphi_\alpha= \{ 0 \ 0 \ 1 \ 0 \ 0 \}$.

\subsection{Learning for Individual Feature}

In the second learning step, the objective is to identify the promising features through cognitive and social learning. The second learning set, `$\psi$', is also an $n$-dimensional string which is derived by identifying all the features that are not included in the particle's current position $x$, but are present in the learning exemplars such as $pbest$ and $gbest$. 

For this purpose, consider generic $n$-dimensional binary strings, `$\alpha$ and $\beta$', where the former represents any learning exemplar ($pbest, gbest$) and the later represents the position of the particle, $x$.  

In order to derive the second learning set $\psi_\alpha$ (which represents the learning set derived from $\alpha$ and $\beta$), a bit wise \emph{logical `AND'} operation is performed on $\alpha$ and $\beta$ as follows: 
\begin{linenomath*}
\begin{align}
\label{eq:psi}
  \psi_\alpha = \alpha \wedge \overline{\beta}
\end{align}
\end{linenomath*}
where, $\overline{\beta}$ is the complement of $\beta$ and `$\wedge$' represents the bit wise logical `AND' operator.

Apart from the learning acquired through cognitive thinking  and social interactions, it is equally important to adjust the velocity based on the particle position. This can be achieved by adjusting the selection probability of the features included in the position. For this purpose, the position `$x$' of the particle is used as the \emph{self learning} set.

Consider a learning exemplar, $\alpha= \{ 1 \ 0 \ 1 \ 1 \ 0 \}$ and a particle, $\beta=\{ 1 \ 1 \ 0 \ 0 \ 1 \}$. Following the proposed approach the learning from the exemplar `$\psi_\alpha$' and particle's self learning `$\psi_\beta$' can be derived as follows:\begin{linenomath*}
\begin{align*}
    \psi_\alpha & = \alpha \wedge \overline{\beta} \Rightarrow \{ 1 \ 0 \ 1 \ 1 \ 0 \} \wedge \{ 0 \ 0 \ 1 \ 1 \ 0 \} =\{ 0 \ 0 \ 1 \ 1 \ 0 \}\\
    \psi_\beta & =\beta =\{ 1 \ 1 \ 0 \ 0 \ 1 \}
\end{align*}
\end{linenomath*}
In the final stage of the learning process, the final learning set `$L$' is derived by combining the cardinality and the feature learning sets into a single matrix, as follows:\begin{linenomath*}
\begin{align}
\label{eq:finalls}
    L = \begin{bmatrix}  \varphi & \psi  \end{bmatrix} ^ T
\end{align}
\end{linenomath*}                               
The final learning set $L_\alpha$ is obtained using above procedure and is given as,
\begin{align*}
    L_\alpha = \begin{bmatrix}  \varphi_\alpha \\
                                \psi_\alpha  \end{bmatrix} = \begin{bmatrix} 0 & 0 & 1 & 0 & 0  \\
                                0 & 0 & 1 & 1 & 0 \end{bmatrix}
\end{align*}
Note that, the subscript $\alpha$ in $L$,$\varphi$ and $\psi$ represents the learning sets corresponding to string $\alpha$.

\subsection{Velocity Update}

\begin{algorithm}[t]
    \small
    \SetKwInOut{Input}{Input}
    \SetKwInOut{Output}{Output}
    \SetKwComment{Comment}{*/ \ \ \ }{}
    \Input{$v_i$, $x_i$, $pbest_i$ and $gbest$}
    \Output{$v_i$}
    \BlankLine
    \Comment*[h] {Learning for Subset Cardinality}\\
    Determine the cardinality of the exemplars ($pbest_i$,$gbest$) and the particle position ($x_i$) \\
    Evaluate the cardinality learning sets: $\varphi_{cog}$, $\varphi_{soc}$ and $\varphi_{self}$
    \BlankLine
    \Comment*[h] {Learning for Features}\\   
    Evaluate Feature Learning Sets: $\psi_{cog}$ and $\psi_{soc}$ using (\ref{eq:psi}),  $\psi_{self}=x_i$ \\
    Evaluate the final learning sets: $L_{cog}$, $L_{soc}$ and $L_{self}$ using (\ref{eq:finalls})
    \BlankLine
    \Comment*[h] {Velocity Update}\\  
    Evaluate the influence of the self-learning ($\Delta_i$) using (\ref{eq:delta}) and (\ref{eq:fitfeedback})\\
    Update velocity ($v_i$) using (\ref{eq:velprop})
\caption{2-D learning approach to the velocity update of the $i^{th}$ particle}
\label{fig:velprop}
\end{algorithm}

The next step in the learning process is to control the influence of the learning sets derived from the exemplars and through self-learning. The influence of the cognitive and social learning sets is controlled through acceleration constants $c_1$ and $c_2$, similar to (\ref{eq:stdpso}). 

This approach can be extended to control the influence of \emph{self-learning} derived from the particle's current position. However, it would be more effective to embed the performance (fitness or criterion function) corresponding to the position for the influence control. For example, the particle with comparatively better performance than its peers should exert more influence, \textit{i.e.}, the selection probability of the features included in its position should receive comparatively higher increment. This is achieved by the following function,
\begin{linenomath*}
\begin{gather}
\label{eq:delta}
        \delta_i = 1- \frac{f_{i}^{t}}{max(F^{t})}
\end{gather}
\end{linenomath*}
where, $``f_{i}^{t}"$ is the fitness of the $i^{th}$ particle at iteration $t$ and $``F^t"$ is the vector containing the fitness of the entire swarm at iteration $t$. For each of the particle, the change in probability, $\delta$, is limited between $[0,1]$ and it is evaluated on the basis of its relative performance through (\ref{eq:delta}). Further, the change in probability ($\delta$) is positive only when the position is deemed beneficial, \textit{i.e.}, $f_{i}^{t}<f_{i}^{t-1}$ (for a minimization problem). For this purpose, the following function $``\Delta"$ for the $i^{th}$ particle is used.
\begin{linenomath*} 
\begin{gather}
\label{eq:fitfeedback}
       \Delta_i= \begin{cases}
                + \delta_i, &   if \; \; \frac{f_{i}^{t}}{f_{i}^{t-1}}<1\\
                - \delta_i, &   otherwise
                 \end{cases}
\end{gather}
\end{linenomath*}

Note that the function ``$\Delta_i$" is similar to the function used to evaluate the self contribution of the $i^{th}$ particle to the velocity update for the assignment problems in \cite{Hafiz:Abdennour:2016}. Finally, the velocity update based on the proposed approach is given by,
\begin{linenomath*}
\begin{align}
\label{eq:velprop}
    v_i= (\omega \times v_i) + (c_1r_1 \times L_{cog}) + (c_2r_2 \times L_{soc}) + (\Delta_i \times L_{self})
\end{align}
\end{linenomath*}
where, $L_{cog}$ is the \emph{cognitive} learning set derived from $pbest$, $L_{soc}$ is the \emph{social} learning set derived from $gbest$ and $L_{self}$ is the \emph{self} learning derived from the particle position, $x_i$.

In the proposed approach the velocity is not bounded, that is, there is no need to specify and adjust $[v_{min},v_{max}]$, which has significant influence on the exploration/exploitation trade-off in the other binary versions of PSO \cite{Kennedy:Eberhart:1997}. The pseudo code for the proposed approach to velocity update is shown in Algorithm-\ref{fig:velprop}.

\subsection{Illustrative Example}
\label{sec:example}
Consider a dataset having five number of features \textit{i.e.}, $n=5$. Let the position, the velocity and the learning exemplars of the $i^{th}$ particle be,
\begin{linenomath*} 
\begin{align*}
    x_i & =\{ 1 \ 0 \ 1 \ 0 \ 1 \}, \ pbest_i= \{ 0 \ 1 \ 0 \ 0 \ 1 \}, \ gbest = \{ 1 \ 1 \ 0 \ 1 \ 0 \}, \ v_i & =\begin{bmatrix} p_{11}^i & p_{12}^i & p_{13}^i & p_{14}^i & p_{15}^i \\ 
                        p_{21}^i & p_{22}^i & p_{23}^i & p_{24}^i & p_{25}^i \end{bmatrix}
\end{align*}
\end{linenomath*}
Based on the proposed learning approach (as shown in Algorithm-\ref{fig:velprop}), the velocity of the $i^{th}$ particle is updated as follows:
\begin{enumerate}
    \item Determination of  the cardinality of all sets ($\xi_{cog}, \xi_{soc}$ and $\xi_{self}$):
    \begin{linenomath*}
    \begin{align*}
            pbest_{i} = \{0 \, 1 \, 0 \, 0 \, 1\} \Rightarrow \xi_{cog} = 2, 
            \ gbest = \{1 \, 1 \, 0 \, 1 \, 0\} \Rightarrow \xi_{soc} = 3, \ x_{i} = \{1 \, 0 \, 1 \, 0 \, 1\} \Rightarrow \xi_{self} = 3.
    \end{align*}
   \end{linenomath*}
    \item Evaluation of the cardinality learning sets ($\varphi_{cog}, \varphi_{soc}$, and $\varphi_{self}$):
    \begin{linenomath*} 
        \begin{align*}
            \xi_{cog}=2 \Rightarrow \varphi_{cog}= \{ 0 \, 1 \, 0 \, 0 \, 0 \}, \ \xi_{soc}=3 & \Rightarrow \varphi_{soc}=\{ 0 \, 0 \, 1 \, 0 \, 0 \}, \ \xi_{self}=3 & \Rightarrow \varphi_{self}=\{ 0 \, 0 \, 1 \, 0 \, 0 \}.
        \end{align*}
    \end{linenomath*}
    \item Evaluation of the feature learning sets($\psi_{cog}, \psi_{soc}$, and $\psi_{self}$):
    \begin{linenomath*} 
            \begin{align*}
                \psi_{cog} = pbest_i \wedge \overline{x_i} = \{ 0 \ 1 \ 0 \ 0 \ 0 \}, \ \psi_{soc} = gbest \wedge \overline{x_i} = \{ 0 \ 1 \ 0 \ 1 \ 0 \}, \ \psi_{self} =x_i=\{ 1 \ 0 \ 1 \ 0 \ 1 \}.
            \end{align*}
    \end{linenomath*}
    \item Derivation of the final learning sets ($L_{cog}, L_{soc}$, and $L_{self}$)
    \begin{linenomath*} 
        \begin{align*}
           L_{cog} & = \begin{bmatrix} \varphi_{cog} & \psi_{cog} \end{bmatrix}^T &   \Rightarrow L_{cog} = \begin{bmatrix} 0 & 1 & 0 & 0 & 0\\
                                                                                              0 & 1 & 0 & 0 & 0 \end{bmatrix} \\
            L_{soc} & = \begin{bmatrix} \varphi_{soc} & \psi_{soc} \end{bmatrix}^T    &   \Rightarrow L_{soc} = \begin{bmatrix} 0 & 0 & 1 & 0 & 0\\
                                                                                              0 & 1 & 0 & 1 & 0 \end{bmatrix}\\
            L_{self} & =\begin{bmatrix} \varphi_{self} & \psi_{self} \end{bmatrix}^T &   \Rightarrow L_{self} = \begin{bmatrix} 0 & 0 & 1 & 0 & 0\\
                                                                                               1 & 0 & 1 & 0 & 1 \end{bmatrix}
        \end{align*}
    \end{linenomath*}
    \item  Velocity update of the $i^{th}$ particle ($v_i$) as per (\ref{eq:velprop})
        \begin{linenomath*} 
        \begin{align}
        \small
        \label{eq:velupdate}
            v_i=\begin{bmatrix} (\omega \times p_{11}^i) & (\omega \times p_{12}^i)+c_1r_1 & (\omega \times p_{13}^i)+c_2r_2+\Delta_i   & (\omega \times p_{14}^i) & (\omega \times p_{15}^i) \\ 
             (\omega \times p_{21}^i)+\Delta_i & (\omega \times p_{22}^i)+c_1r_1+c_2r_2 & (\omega \times p_{23}^i)+\Delta_i & (\omega \times p_{24}^i)+c_2r_2 & (\omega \times p_{25}^i)+\Delta_i \end{bmatrix}
        \end{align}
        \end{linenomath*}
\end{enumerate}

\subsection{Position Update}
\label{sec:position}
In the majority of binary PSO variants, the position update process entails the decision to exclude or include each feature. The probability of inclusion is supplied by the velocity. The major drawback of this approach is the inability to exclude redundant features, as pointed out earlier in the Section \ref{sec:PSOreview}. This is probably due to emphasis given to the significance of the feature without considering about the subset cardinality. To overcome these drawbacks, in this study, additional dimension has been included which stores information about subset cardinality. Consequently, the position update requires the selection of subset cardinality ($\xi$) followed by the selection of features. 

For the example considered in the Section \ref{sec:example}, the velocity of the $i^{th}$ particle is updated using (\ref{eq:velupdate}) and is given by 
\begin{linenomath*} 
\begin{align*}
    v_i=\begin{bmatrix} 0.14 & 2.56 & 1.35 & 0.38 & 0.71\\
                        1.31 & 2.40 & 0.57 & 1.46 & 1.30 \end{bmatrix}
\end{align*}
\end{linenomath*}
For sake of clarity, we introduce to vectors `$\rho$' and $\sigma$ to represent the selection likelihood of cardinality and features,respectively. where,
\begin{linenomath*} 
\begin{align}
\label{eq:rho}
\rho = p_{1,j}^{ \ i}, \ \ j=1, 2, \dots n. \\
\label{eq:sigma}
\sigma = p_{2,j}^{ \ i}, \ \ j=1, 2, \dots n.
\end{align}
\end{linenomath*}

For the example considered here, 
\begin{linenomath*} 
\begin{align}
\label{eq:rho1}
     \rho =\begin{bmatrix} 0.14 & 2.56 & 1.35 & 0.38 & 0.71 \end{bmatrix} \ \text{and} \     \sigma =\begin{bmatrix} 1.31 & 2.40 & 0.57 & 1.46 & 1.30 \end{bmatrix}
\end{align}
\end{linenomath*}
\begin{algorithm}[t]
    \small
    \SetKwInOut{Input}{Input}
    \SetKwInOut{Output}{Output}
    \SetKwComment{Comment}{*/ \ \ \ }{}
    \Input{$v_i$}
    \Output{$x_i$}
    \BlankLine
    Set the new position $x_i$ to a null vector of dimension $n$, \textit{i.e.}, $x_i=\{ 0 \dots 0 \}$ \\
    Isolate the selection likelihood of the \emph{cardinality} and \emph{feature} into respective vectors, `$\rho$' and `$\sigma$' using (\ref{eq:rho}) and (\ref{eq:sigma})
    \BlankLine
    \Comment*[h] {Roulette Wheel Selection of the Subset Cardinality, ($\xi_i$)}\\
    Evaluate accumulative probabilities, $\rho_{\Sigma,j} = \sum \limits_{k=1}^{j} \rho_{k},$ for j = 1\dots n. \nllabel{line:pos1}\\
    Generate a random number, $r \in [0,\rho_{\Sigma,n}]$. \nllabel{line:pos2}\\
    Determine $j\in[1,n]$ such that $\rho_{\Sigma,j-1}<r<\rho_{\Sigma,j}$, this gives the size of the subset $\xi_i$, \textit{i.e.}, $\xi_i=j$. \nllabel{line:pos3}\\
    \BlankLine
    \Comment*[h]{Selection of the features}\\
    Rank the features on the basis of their \emph{likelihood} `$\sigma_j$' and store the feature rankings in vector `$\tau$' \nllabel{line:fs1}\\
    \For{j = 1 to n} 
        { \If{$\tau_{j} \leq \xi_i$}
            {$x_{i,j}=1$}
        } \nllabel{line:fs2}
\caption{2-D learning approach to the position update of the $i^{th}$ particle}
\label{fig:posprop}
\end{algorithm}
There are several ways to select the cardinality ($\xi_i$) of the $i^{th}$ particle. The simplest approach is to select cardinality corresponding to the element with the highest likelihood in $\rho$. For example, in (\ref{eq:rho1}) the second element ($\rho_{2}$) has the highest probability ($2.56$). Hence, in the new position two features ($\xi_i=2$) can be selected. This approach is more \emph{exploitative} in nature, as the size of the subset will always be selected based on the element with highest probability. 

Therefore, in this study an alternate approach based on \emph{roulette wheel} has been used for the selection of the subset cardinality to balance the  the \emph{exploration} and \emph{exploitation} of the search space. Note that in GA, the slot of the \textit{roulette wheel} is assigned to each \emph{parent} based on its \emph{fitness}, whereas in this study the slot is assigned to each \emph{cardinality} on the basis of its \emph{selection likelihood}, as illustrated in Algorithm-\ref{fig:posprop} (Line \ref{line:pos1}-\ref{line:pos3}). 

For example, for the $i^{th}$ particle, the accumulative likelihood (denoted as `$\rho_\Sigma$') is derived from the cardinality likelihood $\rho$ (\ref{eq:rho}) as follows:  
\begin{align}
    \rho_\Sigma= \begin{bmatrix} 0.14 & 2.70 & 4.05 & 4.43 & 5.14 \end{bmatrix}
\end{align}
Assuming the random number, $r \in [0,\rho_{\Sigma,n}]$ is equal to 3.25. Since $\rho_{\Sigma,2}<r<\rho_{\Sigma,3}$, the cardinality of the $i^{th}$ particle is set to 3 ($\xi_i=3$). 

The next step is to select the features to fill the subset. For this purpose, all features are \emph{ranked} on the basis of their \emph{likelihood} ($\sigma_j$) and stored in a separate vector, `$\tau$'. Following this step, the features of the $i^{th}$ particle are ranked as follows:
\begin{align}
   \sigma =\begin{bmatrix} 1.31 & 2.40 & 0.57 & 1.46 & 1.30 \end{bmatrix} \ \
   \Rightarrow \ \ \tau = \begin{bmatrix} 3 & 1 & 5 & 2 & 4\end{bmatrix}
\end{align}
In the final step, the features are selected based on their ranking; all the features with ranking \emph{less than} or \emph{equal to} the selected cardinality $\xi$ are included in the new position, as shown in Algorithm-\ref{fig:posprop} (Line \ref{line:fs1}-\ref{line:fs2}). As per these steps, the new position of the $i^{th}$ particle is derived as follows:
\begin{align}
     \text{since $\xi_i=3$ and \ } \tau = \begin{bmatrix} 3 & 1 & 5 & 2 & 4\end{bmatrix} \Rightarrow x_i=\begin{bmatrix} 1 & 1 & 0 & 1 & 0 \end{bmatrix} \nonumber
\end{align}
\subsection{Refresh Gap}

To overcome the problem of premature convergence, the method of ``\textit{restart}" \cite{Hafiz:Abdennour:2016} is being used in this study to rejuvenate the search process. For this purpose, the personal best value ($pbestval$) of each particle is monitored. If the personal best of any particle does not improve for a pre-determined number of iterations, referred here as `\textit{Refresh Gap}' (RG), its velocity is randomly initialized between $[0,1]$. The pseudo code of GPSO adapted according to 2D-learning framework is shown in Algorithm-\ref{fig:gpso}.
\begin{algorithm}[!t]
    \small
    \SetKwInOut{Input}{Input}
    \SetKwInOut{Output}{Output}
    \SetKwComment{Comment}{*/ \ \ \ }{}
    \Input{Dataset with $n$-number of features}
    \Output{Feature Subset $x$, $\xi<n$}
    \BlankLine
    Set the search parameters: $c_1, \ c_2, \  \omega_0, \  \omega_f \  \& \  RG$ \\
    Randomly initialize the swarm of `$ps$' number of particles, $X=\{ x_1 \dots x_{ps} \}$ \\
    Initialize the velocity ($(2 \times n)$ matrix) of each particle by uniformly distributed random numbers in [0,1] \\
    Evaluate the fitness of the swarm, $pbest$ and $gbest$ \\
    \For{t = 1 to iterations}
        { 
            Evaluate Inertia Weight, $\omega(t) = \frac{(\omega_0 - \omega_f) \times t}{Max \ Iteration}$ \\
            \BlankLine
            \Comment*[h]{Swarm Update}\\
            \For{i = 1 to ps} 
            {
                \BlankLine
                \Comment*[h]{Stagnation Check}\\
                \If{$ count_i \geq RG$}
                    {Re-initialize the velocity of the particle\\
                    Set $count_i$ to zero}
                \BlankLine
                Update the velocity of the $i^{th}$ particle following Algorithm-\ref{fig:velprop} \\
                Update the position of the $i^{th}$ particle following Algorithm-\ref{fig:posprop}
            }
            \BlankLine
            Store the old fitness of the swarm in `$F$'\\
            Evaluate the swarm fitness\\
            Update personnel and global best position, $pbest$, $gbest$\\             
            \BlankLine
            \Comment*[h]{Stagnation Check}\\ 
            \For{i = 1 to ps}
                { \If{$pbestval_i^{ \ t} \geq pbestval_i^{ \ t-1}$}
                        {$count_i=count_i+1$}
                }
        
        }
\caption{Pseudo code of generic GPSO \cite{Kennedy:Eberhart:1995} adapted to feature selection problem on the basis of 2-D learning approach}
\label{fig:gpso}
\end{algorithm}
\section{2-D learning framework for other PSO variants}
\label{sec:adapt}
The proposed learning framework is designed as a generalized learning framework to adapt the algorithm based on particle swarm theory, \textit{i.e.}, The majority of the PSO variants in the continuous domain can easily be adapted for the feature selection problems using the 2-D learning. 

For example, the \emph{local} version of PSO (LPSO) \cite{Kennedy:Mendes:2002}, employs best particle in the particle's neighborhood ($nbest$) as the social learning exemplar instead of $gbest$, as revealed in its velocity update rule, 
\begin{align}
 v_{i} = (\omega \times v_{i}) + c_1r_1 \times (pbest_{i} - x_{i}) + c_2r_2 \times (nbest_{i}-x_{i})    \nonumber
\end{align}
To adapt LPSO for the feature selection problem as per 2-D learning, the social learning set ($L_{soc}$) in (\ref{eq:velprop}) is derived from $nbest_i$ instead of $gbest$.

Another popular PSO variant, \emph{Comprehensive Learning PSO} (CLPSO) \cite{Liang:Qin:2006}, uses only one learning exemplar (referred as $pbest_f$) instead of two, cognitive and social exemplar. Following 2-D learning, the velocity update rule for the adapted CLPSO is given by,
\begin{align}
     v_i= (\omega \times v_i) + (cr \times L) + (\Delta_i \times L_{self})
\end{align}
where, `$L$' is the learning set derived from `$pbest_{f}$'.

The results of our previous study \cite{Hafiz:Abdennour:2016} suggest that, \emph{Unified Particle Swarm Optimization} (UPSO) (\cite{UPSO1}) is quite effective for the problems in the discrete domain. For this reason, UPSO is selected for further analysis in this work. UPSO combines the \textit{global} and \textit{local} versions of PSO through unification factor, `$u$'. The velocity update of UPSO is given by,
\begin{align}
\label{eq:UPSOorig} 
v_{i}  & = (u \times v_{gi})+((1-u) \times v_{li}) \\
\text{where,} \nonumber \\
v_{gi} & = (\omega \times v_{i}) + c_1r_1 \times (pbest_{i} - x_{i}) + c_2r_2 \times (gbest-x_{i}) \nonumber \\ 
v_{li} & = (\omega \times v_{i}) + c_1r_1 \times (pbest_{i} - x_{i}) + c_2r_2 \times (nbest_{i}-x_{i}) \nonumber
\end{align}
and ``$nbest_i$" is the best particle in the neighborhood of the $i^{th}$ particle.

By simple modifications in `$v_{gi}$' and `$v_{li}$', UPSO can be adapted for feature selection problem,
\begin{align}
\label{eq:UPSOupdated1}
v_{gi} & = (\omega \times v_{i}) + (c_1r_1 \times L_{cog}) + (c_2r_2 \times L_{soc,1}) + (\Delta_{i} \times L_{self}) \\ 
\label{eq:UPSOupdated2}
v_{li} & = (\omega \times v_{i}) + (c_1r_1 \times L_{cog}) + (c_2r_2 \times L_{soc,2}) + (\Delta_{i} \times L_{self})
\end{align}
where, `$L_{soc,1}$' and `$L_{soc,2}$' are the social learning sets derived from global best ($gbest$) and neighborhood best ($nbest_i$). The pseudo code for velocity update of the adapted UPSO (denoted as 2D-UPSO) is illustrated in Algorithm-\ref{fig:velupso}. Note that apart from velocity update, rest of the procedure is similar to GPSO as outlined in Algorithm-\ref{fig:gpso}.

\begin{algorithm}[!t]
    \small
    \SetKwInOut{Input}{Input}
    \SetKwInOut{Output}{Output}
    \Input{$v_i$, $x_i$, $pbest_i$ and $gbest$}
    \Output{$v_i$}
    \BlankLine
    \begin{enumerate}
        \item \emph{Locate the best particle in the neighborhood of the $i^{th}$ particle and set it as the second learning exemplar, `$nbest_i$'}
        \item \emph{Complete Learning} 
            \begin{enumerate}
                \item Evaluate Cardinality Learning Sets:\\
                $\varphi_{cog}$, $\varphi_{soc-1}$, $\varphi_{soc-2}$ and $\varphi_{self}$
                \item Evaluate Feature Learning Sets:\\
                $\psi_{cog}$, $\psi_{soc-1}$ and $\psi_{soc-2}$ using (\ref{eq:psi})\\ $\psi_{self}=x_i$
                \item Evaluate the final learning sets:\\
                $L_{cog}$, $L_{self}$, $L_{soc-1}$ and $L_{soc-2}$, using (\ref{eq:finalls})
            \end{enumerate}
        \item \emph{Evaluate the influence of the self-learning  ($\Delta_i$) using (\ref{eq:delta}) and (\ref{eq:fitfeedback})} 
        \item \emph{Update Velocity, $v_i$ using (\ref{eq:UPSOorig}), (\ref{eq:UPSOupdated1}) and (\ref{eq:UPSOupdated2})}
    \end{enumerate}  
\caption{The velocity update of the $i^{th}$ particle in 2D-UPSO}
\label{fig:velupso}
\end{algorithm}
\section{Comparative Evaluation: Benchmarking}
\label{sec:setup}
\subsection{Experimental Setup}

The performance of the proposed 2-D learning framework is compared with the some of the existing algorithms following the \textit{wrapper} approach \cite{Guyon:Isabelle:2003} for the feature selection. Hence, it is necessary to induce a classifier for each particle (PSO based algorithms) or parent (GA) and use the consequent classification performance as criterion function, $J$ in (\ref{eq:fsp}). Without loss of generality, the feature selection problem is modelled as a minimization problem with the objective to \emph{minimize the classification error}. In this work, the mean classification error after \textit{10-fold stratified cross validation} \cite{Witten:Frank:Hall:2016} is being used as the criterion function, \textit{i.e.} for a subset given by a particle/parent ($x_i$), it is given by, 
\begin{linenomath*} 
\begin{align}
\label{eq:fitness}
& J(x_i) = \frac{\sum \limits_{f=1}^{10} e_f}{10}, \text{ \ \ where, \ } e_f=\frac{\text{Number of Misclassified of Samples}}{\text{Total Number of Samples}}
\end{align}
\end{linenomath*}
The use of 10-fold cross validation often removes any bias in the validation data \cite{Witten:Frank:Hall:2016}.

One of the major factor which influences the effectiveness of the feature subset is the nature of the induction algorithm used to induce a classifier. For a given dataset, an optimal feature subset, obtained using a particular classifier, may provide suboptimal performance for the other classifier. In other words, the landscape of the feature selection problem is jointly defined by the dataset and the induction algorithm. For this reason, the efficacy of all the algorithms was evaluated using two induction hypotheses; Naive Bayes (NB) and k-Nearest Neighbor (k-NN) \cite{Witten:Frank:Hall:2016} due to their simplicity and wide scale use. For a k-NN based classifier, number of neighbors in this study were chosen to be 5 ($k=5$).

All the compared algorithms were implemented in MATLAB, R2015b. Due to stochastic nature of the search algorithms, 40 independent runs of each algorithm were performed on each dataset as listed in Table \ref{t:dataset} (More detail on the datasets is provided in the next subsection). In each run, the best feature subset, its cardinality and its fitness were recorded along with the total execution time. For fair comparison, each algorithm was allowed to run for $6000$ \emph{Function Evaluations} (FEs) on each dataset. The process is repeated for both NB and k-NN classifier. 

\subsection{Data Sets}
The efficacy of the proposed learning approach and the other compared algorithms was evaluated using a test bench consisting of 14 data-sets from the UCI Machine Learning Repository \cite{UCI:2013}. The selected datasets, listed in Table \ref{t:dataset}, are representative of different application areas (Financial, Bio-medical and Physical Sciences). In addition, the number of features varies from 13 to 591, output classes vary from 2 to 16 and training instances vary from 27 to 1847. Among sixteen datasets, two datasets, namely `LSVT' and `Lung Cancer' have  higher number of input features ($n$) than number of samples ($\tilde{N}$), ($n>\tilde{N}$) and are \emph{ill-defined}. As shown in Section \ref{sec:res}, the proposed learning framework performs well on these problems as well.

In addition to the datasets, Table \ref{t:dataset} also lists the mean classification error after 10-fold stratified cross validation with NB and 5-NN classifier (k-NN classifier with five neighbors), when all available features are used. These classification errors were used as the reference to evaluate percentage improvement in the classification error obtained with the reduced subsets. The results are discussed at length in Section \ref{sec:res}.
\begin{table}[!t]
\centering
\begin{adjustbox}{max width=0.7\textwidth}
\begin{threeparttable}
\caption{Datasets included in the study from the UCI Machine Learning Repository \cite{UCI:2013}}
\label{t:dataset} 
\centering
\small
\begin{tabular}{c c c c c c} 
 \toprule
 {\textbf{Dataset}} & {\textbf{Features ($n$)}} & {\textbf{Instances ($\tilde{N}$)}} & {\textbf{Output Class}} & {\textbf{$J(U)$ (NB)}} & {\textbf{$J(U)$ (k-NN)}}\\
  \midrule
    Wine	&	13	&	178	&	3	&	0.6007	&	0.3157	\\
    Australian	&	14	&	690	&	2	&	0.3217	&	0.3739	\\
    Zoo	&	17	&	101	&	7	&	0.4121	&	0.4825	\\
    Vehicle	&	18	&	846	&	4	&	0.7234	&	0.6311	\\
    German	&	24	&	1000	&	2	&	0.3120	&	0.3660	\\
    WDBC	&	30	&	569	&	2	&	0.3726	&	0.1300	\\
    Ionosphere	&	34	&	351	&	2	&	0.6790	&	0.1081	\\
    Lung Cancer$^{\dagger}$	&	56	&	27	&	3	&	0.5500	&	0.4000	\\
    Sonar	&	60	&	208	&	2	&	0.5333	&	0.4717	\\
    Ozone	&	72	&	1847	&	2	&	0.0693	&	0.0709	\\
    Libras	&	90	&	360	&	15	&	0.9310	&	0.8470	\\
    Hill	&	100	&	606	&	2	&	0.4819	&	0.4951	\\
    Musk-1	&	166	&	476	&	2	&	0.4348	&	0.4724	\\
    Arrhythmia$^{\dagger}$ 	&	279	&	452	&	16	&	0.8538	&	0.3876	\\
    LSVT	&	310	&	126	&	2	&	0.6110	&	0.1965	\\
    SECOM$^{\dagger}$	&	591	&	1567	&	2	&	0.0817	&	0.0689	\\
 \bottomrule
\end{tabular}
\begin{tablenotes}
      \scriptsize
      \item $^{\dagger}$ - Missing data were imputed using Nearest Neighbor approach 
    \end{tablenotes}
  \end{threeparttable}
\end{adjustbox}
\end{table}
\begin{table}[!t]
\centering
\small
\begin{adjustbox}{max width=0.7\textwidth}
\begin{threeparttable}
\caption{Search Parameter Settings}
\label{t:sparam} 
\begin{tabular}{c c c c c c} 
 \toprule
  \multirow{3}{*}{\textbf{Algorithm}} &  \multicolumn{2}{c} {\textbf{Search Parameters}}  \\
   & \textbf{General PSO Parameters} & \multirow{2}{*}{\textbf{Other/Special Parameters}} \\
   & $[ps, \ \omega, \ c_1, \ c_2, \ v_{min}, \ v_{max} ]$  &   \\
 \midrule
GA \cite{Siedlecki:Sklansky:1989,Grefenstette:1986} &  - &  $N=80$, $p_c=0.45$, $p_m=0.01$\\
ACO \cite{Yu:Gu:2009} &  - &  $as=50$, $a=5$, $\rho=0.2$, $[\tau_{min},\tau_{max}]=[0.3, 1.5]$\\
BPSO \cite{Kennedy:Eberhart:1997}                   &  $[30, \ 1, \ 2, \ 2, \ -6, \ 6]$ & - \\
CBPSO \cite{Chuang:Tsai:2011}                &  $[30, \ 1, \ 2, \ 2, \ -6, \ 6]$ & $C=\frac{ps}{10}$, RG=3 \\
ErFS \cite{Lin:Ying:2008,Xue:Zhang:2013}            & $[30, \ 0.729, \ 1.49, \ 1.49, \ -, \ -]$ & $\theta=0.6$, $[x_{min},x_{max}]=[0,1]$\\
PSO-42 \cite{Xue:Zhang:2014}                        & $[30, \ 0.729, \ 1.49, \ 1.49, \ -, \ -]$ & $\theta=0.6$, $[x_{min},x_{max}]=[0,1]$\\
chBPSO \cite{Chuang:Yang:2011}                     &  $[30, -, \ 2, \ 2, \ -6, \ 6]$ & variable $\omega$ \\
2D-UPSO (proposed)                                    &  $[30, \ 0.729, \ 1.49, \ 1.49, \ -, \ -]$ & $u:0.2 \rightarrow 0.4$, RG=3\\
\bottomrule
\end{tabular}
\begin{tablenotes}
      \scriptsize
      \item `$ps$' - swarm size, `$\omega$' - inertia weight, `$c_1,c_2$' - acceleration constants, $v_{min},v_{max}$ - velocity limits  
      \item `$N$' - GA population size, $p_c,p_m$ - crossover and mutation probability
      \item `$as$' - colony size, `$a$' - pheromone update factor, `$\rho$' - pheromone trail evaporation, $[\tau_{min},\tau_{max}]$ - pheromone boundaries  
      \item `$C$' - catfish particles, `$\theta$' - threshold to select feature, $x_{min},x_{max}$ - position limits, $u$ - unification factor
    \end{tablenotes}
  \end{threeparttable}
 \end{adjustbox}
\end{table}
\subsection{Compared Algorithms}

To evaluate the potency of the proposed learning approach following seven algorithms were included as benchmarking algorithms in the comparative evaluation: Genetic Algorithm (GA) \cite{Siedlecki:Sklansky:1989}, Ant Colony Optimization (ACO) \cite{Yu:Gu:2009,Chen:Chen:2013}, Binary PSO (BPSO) \cite{Kennedy:Eberhart:1997}, Catfish Binary PSO (CBPSO) \cite{Chuang:Tsai:2011}, Chaotic Binary PSO (chBPSO) \cite{Chuang:Yang:2011}, variant based on continuous PSO (ErFS) \cite{Lin:Ying:2008,Xue:Zhang:2013} and PSO with different update mechanisms for `$gbest$' (PSO-42) \cite{Xue:Zhang:2014}.

The proposed approach is in essence a new learning strategy for PSO hence it would be interesting to benchmark its performance using existing PSO based feature selection approaches. For this purpose, Kennedy and Eberhart's binary version of PSO (BPSO) \cite{Kennedy:Eberhart:1997} is a natural choice, as majority of the PSO based feature selection approaches are either directly based on it or some extension of it. Catfish-BPSO (CBPSO) has a learning mechanism which is similar to BPSO; however it employs the concept of `Refresh Gap' in \cite{Liang:Qin:2006} to replace fraction of worst performing particles if `$gbest$' does not improve for the pre-determined iterations. This measure is in essence a `Restart' approach and is expected to perform better than BPSO. Since, the proposed approach also include restart mechanism, it would be interesting to have comparative analysis with Catfish-BPSO. On the other hand, ErFS \cite{Lin:Ying:2008,Xue:Zhang:2013} and PSO-42 \cite{Xue:Zhang:2014} are extension of continuous PSO and represent different learning approaches and therefore are included as the benchmarking algorithms. The last benchmarking algorithm is Chaotic Binary PSO (chBPSO) \cite{Chuang:Yang:2011}, which is in essence extension of BPSO with an updated control strategy for the inertia weight based on chaotic map theory. Even though its learning approach is similar to BPSO, chBPSO represents the PSO variant with a parameter control and for this reason is included in the benchmarking algorithms. In addition to PSO variants, two other search paradigms, GA and ACO, have been included as benchmarking algorithms. Note that GA is, arguably, the most widely used meta-heuristics algorithm for the feature selection problems and therefore included as a benchmarking algorithm. ACO is selected as another benchmarking algorithm, since its underlying search principle is similar to PSO. 

Note that, all benchmarking algorithms were implemented in MATLAB following the procedure and parameters provided in  \cite{Siedlecki:Sklansky:1989,Yu:Gu:2009,Chen:Chen:2013,Kennedy:Eberhart:1997,Chuang:Tsai:2011,Lin:Ying:2008,Xue:Zhang:2014,Chuang:Yang:2011}. The search parameters for the all algorithms are listed in Table \ref{t:sparam}. Note that even-though, GA has been used widely, there is no common consensus on the good parameter setting, i.e., population size, crossover and mutation rate $(N,p_c,p_m)$. In this work, several preliminary runs of GA were carried out with different parameter settings suggested in the earlier works \cite{Siedlecki:Sklansky:1989,Grefenstette:1986,Yang:Honavar:1998,Raymer:Punch:2000,Sikora:Piramuthu:2007}. In this test the parameter settings suggested by \cite{Grefenstette:1986} were found to be more effective and hence were selected for further analysis. Moreover, Chuang et. al. \cite{Chuang:Yang:2011} suggested two control strategies for the inertia weight: based on logistic map and tent map. In this work, logistic map based approach is being used for the inertia weight control given by,
\begin{linenomath*} 
\begin{equation*}
    \omega^{t+1}=4 \times \omega^t \times (1-\omega^t), \text{ \ where, \ } \omega^0=0.48 \ \text{ and  $t$ is the iteration number} 
\end{equation*}
\end{linenomath*}

The unification factor, `$u$' in (\ref{eq:UPSOorig}), plays a major role in determining the search behavior of UPSO. Several approaches including \textit{linearly increasing}, \textit{modular} and \textit{exponential changes} therefore have been proposed to adapt $u$ during search process \cite{UPSO2}. In the present study, linearly increasing $u$ with iterations (from 0.2 to 0.4) was found to be most effective and hence selected for the analysis.

\section{Results and Discussion}
\label{sec:res}

The objective of the comparative analysis is to evaluate the performance of the proposed learning approach with the benchmarking algorithms. The performance of these algorithms is compared using two metrics: the reduction in the \textit{classification error} and \textit{subset cardinality}. Moreover, the algorithms are applied as wrappers to two classifiers \textit{k-NN} and \textit{NB}. Due to stochastic nature of these algorithms, the performance is compared using the average values of the metrics obtained from 40 independent runs.

\subsection {Improvement in Classification Performance} 

The results of the classification performance for all the compared algorithms on the benchmark datasets (Table \ref{t:dataset}) are given in Table \ref{t:er:nb} for \textit{NB} classifier and in Table \ref{t:er:knn} for \textit{k-NN} classifier. The last columns of these tables show the results obtained by the proposed \textit{2-D learning approach}. 

For each of the dataset, the \textit{average} (Mean) and \textit{standard deviation} (SD) of the \textit{classification error} were obtained after 40 independent runs of each algorithm. Since the main objective is to minimize the classification error, the algorithms were compared using \textit{Performance Improvement} (PI) metric which is defined as:
\begin{equation}
PI (\%)=\frac {J(U)-J(X)}{J(U)} \times 100
\end{equation}
where, $J(U)$ is the classification error obtained with all features ($U=\{ u_1 \dots u_n\}$) and $J(X)$ is the classification error obtained with the reduced subset $X= \{ u_1 \dots u_\xi \}$ which is found by the various algorithms. Further, for each dataset, a \textit{rank} is assigned to each algorithm based on its \textit{Performance Improvement} (PI). For example, in Table \ref{t:er:nb} for `\textit{wine}' dataset, the best \textit{PI} is obtained by chBPSO and ACO ($65.3 \%$), whereas PSO-42 provided worst `PI' ($59.2\%$). Consequently, both chBPSO and ACO are ranked `1' and rank `8' is assigned to PSO-42. 

The results of this experiment corroborate the hypothesis that the search landscape is conjointly defined by the dataset and the induction algorithm. The results shown in Tables \ref{t:er:nb}-\ref{t:er:knn} indicate that, for a given dataset, the classification performance is critically dependent on the nature of the induction algorithm being used to induce the classifier. For example, for `WDBC' dataset, the PI obtained with NB classifier by all the algorithms lies in the range of $[0,7.1 \%]$ (as seen in Table \ref{t:er:nb}), whereas this lies in the range of $[60.4 \%,62.2\%]$ with k-NN classifier (as seen Table \ref{t:er:knn}).

Note that for several datasets, the PI from some of the algorithms is \textit{negative}, as highlighted in \textit{italics} in Tables \ref{t:er:nb} and \ref{t:er:knn}. A negative value of PI indicates that the corresponding algorithm failed to provide a feature subset that can provide better classification performance compared to the original set containing all the features. As seen from Table \ref{t:er:nb}, with NB classifier, the GA failed to provide better accuracy for the `\textit{Ozone}' dataset. Further, for the `\textit{LSVT}' dataset (Table \ref{t:er:knn}), none of the compared algorithm provided the performance improvement over the original dataset, however, the proposed 2D-UPSO, improves the performance by $29.6\%$. Note that for all the datasets and both the classifiers, the value of PI obtained by 2D-UPSO is positive (as seen in Tables \ref{t:er:nb} and \ref{t:er:knn}) which implies that, for all the dataset, 2D-UPSO is able to provide feature subset that can provide enhanced performance over the original set with all the features. 

Among PSO variants, ErFS and PSO-42 performed very poorly, which was expected as these variants are slightly updated version of the continuous PSO and does not include any meaningful learning mechanism. For most of the datasets, both ErFS and PSO-42 performed poorly with the ranks in the range of 4-7, as seen in Tables \ref{t:er:nb} and \ref{t:er:knn}. Only on few datasets, PSO-42 performed better than other chBPSO (Sonar with NB; Musk-1 and Libras with k-NN), BPSO (WDBC with NB) and CBPSO (WDBC with NB; LSVT with k-NN). Nevertheless, on several datasets, PSO-42 outperforms GA (on 11 datasets with both NB and k-NN) and ACO (on 7 datasets with NB and on 5 datasets with k-NN).

To evaluate the \textit{overall} performance of the algorithms, \textit{average rank} over all the datasets was determined for each of the algorithm which forms the basis for the \textit{final rank}, as shown in the last rows of Table \ref{t:er:nb} and \ref{t:er:knn}. From Table \ref{t:er:nb} and \ref{t:er:knn}, it is observed that 2D-UPSO gives the best overall performance with both the induction algorithms; its average rank equals to $1.7$ (with NB classifier) and $2.1$ (with k-NN classifier). The next best overall performance is obtained by BPSO, followed by chBPSO and CBPSO, with both NB and k-NN classifier. The overall performance of GA was inferior compared to the other algorithms, which is probably due to its inability to escape local minima. Note that, interestingly, BPSO outperformed not only GA and ACO but all the compared PSO variants, \textit{i.e.},CBPSO,chBPSO, ErFS and PSO-42. Further, with both the classifiers, ACO achieved overall fifth rank; performing better than GA and only two PSO variants, ErFS and PSO-42. Relatively poor performance of ACO can be ascribed to a common \textit{pheromone trail} based search strategy; in each iteration all ants rely on a common pheromone trail to generate a solution. On the other hand, in PSO, each particle has its own velocity which improves the \textit{`exploration'} of search space and consequently translates into better performance.

\begin{table}
\scriptsize
\centering
\begin{adjustbox}{max width=0.75\textwidth}
\begin{threeparttable}
\caption{Classification performance of the compared algorithms with NB classifier (averaged over 40 runs)}
\label{t:er:nb} 
\centering
\begin{tabular}{c c c c c c c c c c} 
 \toprule
 {\textbf{Dataset}} & {\textbf{Result}} & {\textbf{GA}} & {\textbf{ACO}} & {\textbf{BPSO}} & {\textbf{CBPSO}} & {\textbf{ErFS}} & {\textbf{PSO-42}} & {\textbf{chBPSO}} & {\textbf{2D-UPSO}} \\ [1.0ex] 
 
\midrule \multirow{4}{*}{Wine}	&	{Mean}	&	0.2259	&	0.2085	&	0.2151	&	0.2166	&	0.2419	&	0.2453	&	0.2085	&	0.2122	\\ [0.5 ex]
	&	{SD}	&	1.27E-02	&	5.62E-17	&	8.62E-03	&	9.35E-03	&	3.12E-02	&	3.43E-02	&	5.62E-17	&	6.85E-03	\\ [0.5 ex]
	&	PI (\%)	&	62.4	&	65.3	&	64.2	&	63.9	&	59.7	&	59.2	&	65.3	&	64.7	\\ [0.5 ex]
	&	Rank	&	6	&	1	&	4	&	5	&	7	&	8	&	1	&	3	\\ [0.5 ex]
\midrule \multirow{4}{*}{Australian}	&	{Mean}	&	0.3137	&	0.3117	&	0.3122	&	0.3124	&	0.3132	&	0.3130	&	0.3118	&	0.3118	\\ [0.5 ex]
	&	{SD}	&	1.13E-03	&	5.69E-06	&	9.05E-04	&	1.03E-03	&	1.45E-03	&	1.49E-03	&	2.92E-04	&	2.09E-04	\\ [0.5 ex]
	&	PI (\%)	&	2.5	&	3.1	&	3.0	&	2.9	&	2.6	&	2.7	&	3.1	&	3.1	\\ [0.5 ex]
	&	Rank	&	8	&	1	&	4	&	5	&	7	&	6	&	3	&	2	\\ [0.5 ex]
\midrule \multirow{4}{*}{Zoo}	&	{Mean}	&	0.0016	&	0	&	0	&	0	&	0.0106	&	0.0118	&	0	&	0	\\ [0.5 ex]
	&	{SD}	&	3.55E-03	&	0	&	0	&	0	&	8.63E-03	&	9.38E-03	&	0	&	0	\\ [0.5 ex]
	&	PI (\%)	&	99.6	&	100	&	100	&	100	&	97.4	&	97.1	&	100	&	100	\\ [0.5 ex]
	&	Rank	&	6	&	1	&	1	&	1	&	7	&	8	&	1	&	1	\\ [0.5 ex]
\midrule \multirow{4}{*}{Vehicle}	&	{Mean}	&	0.4517	&	0.4410	&	0.4293	&	0.4290	&	0.4566	&	0.4598	&	0.4278	&	0.4276	\\ [0.5 ex]
	&	{SD}	&	9.95E-03	&	7.59E-03	&	4.74E-03	&	4.56E-03	&	1.78E-02	&	1.58E-02	&	1.64E-03	&	2.26E-03	\\ [0.5 ex]
	&	PI (\%)	&	37.6	&	39.0	&	40.6	&	40.7	&	36.9	&	36.4	&	40.9	&	40.9	\\ [0.5 ex]
	&	Rank	&	6	&	5	&	4	&	3	&	7	&	8	&	2	&	1	\\ [0.5 ex]
\midrule \multirow{4}{*}{German}	&	{Mean}	&	0.2533	&	0.2403	&	0.2402	&	0.2404	&	0.2509	&	0.2509	&	0.2393	&	0.2403	\\ [0.5 ex]
	&	{SD}	&	5.38E-03	&	1.71E-03	&	3.11E-03	&	3.46E-03	&	8.37E-03	&	6.25E-03	&	1.42E-03	&	3.32E-03	\\ [0.5 ex]
	&	PI (\%)	&	18.8	&	23.0	&	23.0	&	23.0	&	19.6	&	19.6	&	23.3	&	23.0	\\ [0.5 ex]
	&	Rank	&	8	&	4	&	2	&	5	&	6	&	7	&	1	&	3	\\ [0.5 ex]
\midrule \multirow{4}{*}{WDBC}	&	{Mean}	&	0.3710	&	0.3689	&	0.3681	&	0.3725	&	0.3723	&	0.3671	&	0.3655	&	0.3463	\\ [0.5 ex]
	&	{SD}	&	5.94E-04	&	3.69E-03	&	8.98E-03	&	3.87E-04	&	6.40E-04	&	8.50E-03	&	6.04E-03	&	3.53E-03	\\ [0.5 ex]
	&	PI (\%)	&	0.4	&	1.0	&	1.2	&	0.0	&	0.1	&	1.5	&	1.9	&	7.1	\\ [0.5 ex]
	&	Rank	&	6	&	5	&	4	&	8	&	7	&	3	&	2	&	1	\\ [0.5 ex]
\midrule \multirow{4}{*}{Ionosphere} 	&	{Mean}	&	0.2571	&	0.1060	&	0.0986	&	0.0983	&	0.1080	&	0.1155	&	0.1121	&	0.0968	\\ [0.5 ex]
	&	{SD}	&	3.66E-02	&	5.31E-03	&	2.27E-03	&	1.90E-03	&	8.99E-03	&	1.85E-02	&	9.23E-03	&	4.50E-04	\\ [0.5 ex]
	&	PI (\%)	&	62.1	&	84.4	&	85.5	&	85.5	&	84.1	&	83.0	&	83.5	&	85.7	\\ [0.5 ex]
	&	Rank	&	8	&	4	&	3	&	2	&	5	&	7	&	6	&	1	\\ [0.5 ex]
\midrule \multirow{4}{*}{Lung Cancer}	&	{Mean}	&	0.2738	&	0.1504	&	0.1240	&	0.1292	&	0.2033	&	0.2017	&	0.1706	&	0.1250	\\ [0.5 ex]
	&	{SD}	&	1.97E-02	&	1.36E-02	&	2.15E-02	&	2.18E-02	&	4.36E-02	&	3.79E-02	&	2.00E-02	&	1.93E-02	\\ [0.5 ex]
	&	PI (\%)	&	50.2	&	72.7	&	77.5	&	76.5	&	63.0	&	63.3	&	69.0	&	77.3	\\ [0.5 ex]
	&	Rank	&	8	&	4	&	1	&	3	&	7	&	6	&	5	&	2	\\ [0.5 ex]
\midrule \multirow{4}{*}{Sonar}	&	{Mean}	&	0.4843	&	0.4034	&	0.2406	&	0.2712	&	0.3747	&	0.3084	&	0.3324	&	0.2540	\\ [0.5 ex]
	&	{SD}	&	2.96E-02	&	2.48E-02	&	1.80E-02	&	7.90E-02	&	6.44E-02	&	2.73E-02	&	1.69E-02	&	1.18E-02	\\ [0.5 ex]
	&	PI (\%)	&	9.2	&	24.4	&	54.9	&	49.2	&	29.7	&	42.2	&	37.7	&	52.4	\\ [0.5 ex]
	&	Rank	&	8	&	7	&	1	&	3	&	6	&	4	&	5	&	2	\\ [0.5 ex]
\midrule \multirow{4}{*}{Ozone}	&	{Mean}	&	0.0717	&	0.0690	&	0.0652	&	0.0656	&	0.0675	&	0.0673	&	0.0666	&	0.0652	\\ [0.5 ex]
	&	{SD}	&	3.35E-03	&	1.01E-03	&	6.91E-04	&	8.96E-04	&	6.78E-04	&	7.22E-04	&	6.45E-04	&	7.31E-04	\\ [0.5 ex]
	&	PI (\%)	&	\textit{-3.5}	&	0.4	&	5.9	&	5.4	&	2.6	&	2.8	&	3.8	&	6.0	\\ [0.5 ex]
	&	Rank	&	8	&	7	&	2	&	3	&	6	&	5	&	4	&	1	\\ [0.5 ex]
\midrule \multirow{4}{*}{Libras}	&	{Mean}	&	0.5722	&	0.5583	&	0.4369	&	0.4391	&	0.5130	&	0.5077	&	0.4961	&	0.4457	\\ [0.5 ex]
	&	{SD}	&	1.30E-02	&	1.05E-02	&	5.67E-03	&	4.48E-03	&	3.15E-02	&	2.83E-02	&	9.98E-03	&	6.88E-03	\\ [0.5 ex]
	&	PI (\%)	&	38.5	&	40.0	&	53.1	&	52.8	&	44.9	&	45.5	&	46.7	&	52.1	\\ [0.5 ex]
	&	Rank	&	8	&	7	&	1	&	2	&	6	&	5	&	4	&	3	\\ [0.5 ex]
\midrule \multirow{4}{*}{Hill}	&	{Mean}	&	0.4690	&	0.4670	&	0.4675	&	0.4675	&	0.4680	&	0.4684	&	0.4670	&	0.4670	\\ [0.5 ex]
	&	{SD}	&	7.55E-04	&	2.59E-04	&	1.14E-03	&	7.76E-04	&	9.71E-04	&	1.11E-03	&	1.69E-16	&	2.59E-04	\\ [0.5 ex]
	&	PI (\%)	&	2.7	&	3.1	&	3.0	&	3.0	&	2.9	&	2.8	&	3.1	&	3.1	\\ [0.5 ex]
	&	Rank	&	8	&	1	&	4	&	5	&	6	&	7	&	2	&	3	\\ [0.5 ex]
\midrule \multirow{4}{*}{Musk-1}	&	{Mean}	&	0.3199	&	0.2738	&	0.2499	&	0.2511	&	0.2887	&	0.2892	&	0.2834	&	0.2188	\\ [0.5 ex]
	&	{SD}	&	7.35E-03	&	4.80E-03	&	6.23E-03	&	9.26E-03	&	1.49E-02	&	1.32E-02	&	5.74E-03	&	1.64E-02	\\ [0.5 ex]
	&	PI (\%)	&	26.4	&	37.0	&	42.5	&	42.3	&	33.6	&	33.5	&	34.8	&	49.7	\\ [0.5 ex]
	&	Rank	&	8	&	4	&	2	&	3	&	6	&	7	&	5	&	1	\\ [0.5 ex]
\midrule \multirow{4}{*}{Arrhythmia}	&	{Mean}	&	0.8106	&	0.8058	&	0.7495	&	0.7346	&	0.7890	&	0.7890	&	0.7900	&	0.3865	\\ [0.5 ex]
	&	{SD}	&	4.86E-03	&	4.48E-03	&	1.15E-02	&	8.35E-02	&	6.59E-03	&	6.80E-03	&	4.40E-03	&	1.95E-02	\\ [0.5 ex]
	&	PI (\%)	&	5.1	&	5.6	&	12.2	&	14.0	&	7.6	&	7.6	&	7.5	&	54.7	\\ [0.5 ex]
	&	Rank	&	8	&	7	&	3	&	2	&	4	&	5	&	6	&	1	\\ [0.5 ex]
\midrule \multirow{4}{*}{LSVT}	&	{Mean}	&	0.1952	&	0.1761	&	0.1272	&	0.1299	&	0.1722	&	0.1695	&	0.1592	&	0.1069	\\ [0.5 ex]
	&	{SD}	&	8.83E-03	&	6.02E-03	&	1.26E-02	&	1.15E-02	&	1.31E-02	&	1.27E-02	&	7.05E-03	&	1.37E-02	\\ [0.5 ex]
	&	PI (\%)	&	68.1	&	71.2	&	79.2	&	78.7	&	71.8	&	72.3	&	73.9	&	82.5	\\ [0.5 ex]
	&	Rank	&	8	&	7	&	2	&	3	&	6	&	5	&	4	&	1	\\ [0.5 ex]
\midrule \multirow{4}{*}{SECOM}	&	{Mean}	&	0.0658	&	0.0648	&	0.0622	&	0.0626	&	0.0646	&	0.0646	&	0.0636	&	0.0621	\\ [0.5 ex]
	&	{SD}	&	9.46E-04	&	4.77E-04	&	9.35E-04	&	8.80E-04	&	7.29E-04	&	7.49E-04	&	3.80E-04	&	4.19E-04	\\ [0.5 ex]
	&	PI (\%)	&	19.4	&	20.7	&	23.9	&	23.3	&	20.8	&	20.9	&	22.1	&	23.9	\\ [0.5 ex]
	&	Rank	&	8	&	7	&	2	&	3	&	6	&	5	&	4	&	1	\\ [0.5 ex]
\midrule \multicolumn{2}{c}{\textbf{Average Rank}}	&			7.5	&	4.5	&	2.5	&	3.5	&	6.2	&	6.0	&	3.4	&	1.7	\\ [0.5 ex]
\multicolumn{2}{c}{\textbf{Overall Rank}}	&			8	&	5	&	2	&	4	&	7	&	6	&	3	&	1	\\ [0.5 ex]
 
\bottomrule
\end{tabular}
 \begin{tablenotes}
      \tiny
      \item `Mean'and 'SD' - Mean and standard deviation of classification error over 40 runs
      \item `PI'- Percentage improvement in the classification error relative to the classification error achieved with all features
    \end{tablenotes}
  \end{threeparttable}
\end{adjustbox}
\end{table}
\begin{table}
\scriptsize
\centering
\begin{adjustbox}{max width=0.75\textwidth}
\begin{threeparttable}
\caption{Classification performance of the compared algorithms with k-NN Classifier (averaged over 40 runs)}
\label{t:er:knn} 
\centering
\begin{tabular}{c c c c c c c c c c} 
 \toprule
 {\textbf{Dataset}} & {\textbf{Result}} & {\textbf{GA}} & {\textbf{ACO}} & {\textbf{BPSO}} & {\textbf{CBPSO}} & {\textbf{ErFS}} & {\textbf{PSO-42}} & {\textbf{chBPSO}} & {\textbf{2D-UPSO}} \\ [1.0ex] 

\midrule \multirow{4}{*}{Wine}	&	{Mean}	&	0.0418	&	0.0340	&	0.0426	&	0.0426	&	0.0450	&	0.0454	&	0.0359	&	0.0384	\\ [0.5 ex]
	&	{SD}	&	5.16E-03	&	0.00E+00	&	4.70E-03	&	4.70E-03	&	1.98E-03	&	2.83E-03	&	4.09E-03	&	5.37E-03	\\ [0.5 ex]
	&	PI (\%)	&	86.8	&	89.2	&	86.5	&	86.5	&	85.8	&	85.6	&	88.6	&	87.8	\\ [0.5 ex]
	&	Rank	&	4	&	1	&	5	&	6	&	7	&	8	&	2	&	3	\\ [0.5 ex]
\midrule \multirow{4}{*}{Australian}	&	{Mean}	&	0.3106	&	0.3044	&	0.3082	&	0.3075	&	0.3123	&	0.3141	&	0.3044	&	0.3045	\\ [0.5 ex]
	&	{SD}	&	5.19E-03	&	6.13E-05	&	5.32E-03	&	5.69E-03	&	6.86E-03	&	8.05E-03	&	6.13E-05	&	9.26E-04	\\ [0.5 ex]
	&	PI (\%)	&	16.9	&	18.6	&	17.6	&	17.8	&	16.5	&	16.0	&	18.6	&	18.5	\\ [0.5 ex]
	&	Rank	&	6	&	2	&	5	&	4	&	7	&	8	&	1	&	3	\\ [0.5 ex]
\midrule \multirow{4}{*}{Zoo}	&	{Mean}	&	0	&	0	&	0	&	0	&	0.0007	&	0.0003	&	0	&	0	\\ [0.5 ex]
	&	{SD}	&	0	&	0	&	0	&	0	&	2.62E-03	&	1.76E-03	&	0	&	0	\\ [0.5 ex]
	&	PI (\%)	&	100.0	&	100	&	100	&	100	&	99.8	&	99.9	&	100	&	100	\\ [0.5 ex]
	&	Rank	&	1	&	1	&	1	&	1	&	8	&	7	&	1	&	1	\\ [0.5 ex]
\midrule \multirow{4}{*}{Vehicle}	&	{Mean}	&	0.2760	&	0.2664	&	0.2698	&	0.2696	&	0.2769	&	0.2765	&	0.2665	&	0.2667	\\ [0.5 ex]
	&	{SD}	&	4.62E-03	&	7.76E-04	&	3.19E-03	&	2.91E-03	&	4.03E-03	&	5.13E-03	&	8.62E-04	&	1.44E-03	\\ [0.5 ex]
	&	PI (\%)	&	56.3	&	57.8	&	57.3	&	57.3	&	56.1	&	56.2	&	57.8	&	57.7	\\ [0.5 ex]
	&	Rank	&	6	&	1	&	5	&	4	&	8	&	7	&	2	&	3	\\ [0.5 ex]
\midrule \multirow{4}{*}{German}	&	{Mean}	&	0.2568	&	0.2478	&	0.2474	&	0.2483	&	0.2575	&	0.2563	&	0.2457	&	0.2472	\\ [0.5 ex]
	&	{SD}	&	4.20E-03	&	2.92E-03	&	4.18E-03	&	4.29E-03	&	5.65E-03	&	6.28E-03	&	2.30E-03	&	4.54E-03	\\ [0.5 ex]
	&	PI (\%)	&	29.8	&	32.3	&	32.4	&	32.2	&	29.7	&	30.0	&	32.9	&	32.5	\\ [0.5 ex]
	&	Rank	&	7	&	4	&	3	&	5	&	8	&	6	&	1	&	2	\\ [0.5 ex]
\midrule \multirow{4}{*}{WDBC}	&	{Mean}	&	0.0493	&	0.0491	&	0.0494	&	0.0493	&	0.0507	&	0.0515	&	0.0491	&	0.0491	\\ [0.5 ex]
	&	{SD}	&	5.67E-04	&	5.62E-17	&	6.51E-04	&	5.67E-04	&	3.06E-03	&	4.32E-03	&	5.62E-17	&	5.62E-17	\\ [0.5 ex]
	&	PI (\%)	&	62.1	&	62.2	&	62.0	&	62.1	&	61.0	&	60.4	&	62.2	&	62.2	\\ [0.5 ex]
	&	Rank	&	4	&	1	&	6	&	5	&	7	&	8	&	1	&	1	\\ [0.5 ex]
\midrule \multirow{4}{*}{Ionosphere} 	&	{Mean}	&	0.0838	&	0.0804	&	0.0697	&	0.0701	&	0.0807	&	0.0807	&	0.0737	&	0.0565	\\ [0.5 ex]
	&	{SD}	&	3.96E-03	&	3.27E-03	&	5.64E-03	&	4.14E-03	&	6.42E-03	&	4.80E-03	&	4.17E-03	&	5.98E-03	\\ [0.5 ex]
	&	PI (\%)	&	22.5	&	25.7	&	35.5	&	35.1	&	25.3	&	25.4	&	31.9	&	47.7	\\ [0.5 ex]
	&	Rank	&	8	&	5	&	2	&	3	&	7	&	6	&	4	&	1	\\ [0.5 ex]
\midrule \multirow{4}{*}{Lung Cancer}	&	{Mean}	&	0.2285	&	0.1390	&	0.1181	&	0.1258	&	0.1800	&	0.1777	&	0.1558	&	0.1381	\\ [0.5 ex]
	&	{SD}	&	2.22E-02	&	1.47E-02	&	2.01E-02	&	2.22E-02	&	3.28E-02	&	3.24E-02	&	2.12E-02	&	1.83E-02	\\ [0.5 ex]
	&	PI (\%)	&	42.9	&	65.3	&	70.5	&	68.5	&	55.0	&	55.6	&	61.0	&	65.5	\\ [0.5 ex]
	&	Rank	&	8	&	4	&	1	&	2	&	7	&	6	&	5	&	3	\\ [0.5 ex]
\midrule \multirow{4}{*}{Sonar}	&	{Mean}	&	0.1383	&	0.1271	&	0.0941	&	0.0954	&	0.1224	&	0.1199	&	0.1123	&	0.0966	\\ [0.5 ex]
	&	{SD}	&	7.22E-03	&	7.49E-03	&	9.56E-03	&	9.28E-03	&	1.06E-02	&	9.75E-03	&	5.80E-03	&	1.08E-02	\\ [0.5 ex]
	&	PI (\%)	&	70.7	&	73.1	&	80.0	&	79.8	&	74.0	&	74.6	&	76.2	&	79.5	\\ [0.5 ex]
	&	Rank	&	8	&	7	&	1	&	2	&	6	&	5	&	4	&	3	\\ [0.5 ex]
\midrule \multirow{4}{*}{Ozone}	&	{Mean}	&	0.0624	&	0.0607	&	0.0568	&	0.0576	&	0.0594	&	0.0593	&	0.0568	&	0.0534	\\ [0.5 ex]
	&	{SD}	&	1.95E-03	&	1.34E-03	&	5.31E-03	&	5.07E-03	&	4.12E-03	&	3.73E-03	&	1.29E-03	&	1.51E-03	\\ [0.5 ex]
	&	PI (\%)	&	12.0	&	14.3	&	19.8	&	18.8	&	16.2	&	16.4	&	19.8	&	24.7	\\ [0.5 ex]
	&	Rank	&	8	&	7	&	2	&	4	&	6	&	5	&	3	&	1	\\ [0.5 ex]
\midrule \multirow{4}{*}{Libras}	&	{Mean}	&	0.2083	&	0.2012	&	0.1671	&	0.1682	&	0.1942	&	0.1929	&	0.1936	&	0.1697	\\ [0.5 ex]
	&	{SD}	&	4.40E-03	&	4.20E-03	&	9.82E-03	&	8.56E-03	&	8.24E-03	&	9.57E-03	&	3.42E-03	&	9.69E-03	\\ [0.5 ex]
	&	PI (\%)	&	75.4	&	76.2	&	80.3	&	80.1	&	77.1	&	77.2	&	77.1	&	80.0	\\ [0.5 ex]
	&	Rank	&	8	&	7	&	1	&	2	&	6	&	4	&	5	&	3	\\ [0.5 ex]
\midrule \multirow{4}{*}{Hill}	&	{Mean}	&	0.4662	&	0.4385	&	0.4158	&	0.4147	&	0.4413	&	0.4421	&	0.4495	&	0.4162	\\ [0.5 ex]
	&	{SD}	&	3.84E-03	&	3.55E-03	&	7.87E-03	&	1.04E-02	&	7.11E-03	&	1.01E-02	&	3.85E-03	&	1.06E-02	\\ [0.5 ex]
	&	PI (\%)	&	5.8	&	11.4	&	16.0	&	16.2	&	10.9	&	10.7	&	9.2	&	15.9	\\ [0.5 ex]
	&	Rank	&	8	&	4	&	2	&	1	&	5	&	6	&	7	&	3	\\ [0.5 ex]
\midrule \multirow{4}{*}{Musk-1}	&	{Mean}	&	0.0946	&	0.0910	&	0.0450	&	0.0460	&	0.0746	&	0.0745	&	0.0803	&	0.0718	\\ [0.5 ex]
	&	{SD}	&	4.68E-03	&	4.01E-03	&	6.86E-03	&	5.54E-03	&	6.26E-03	&	5.44E-03	&	4.53E-03	&	5.51E-03	\\ [0.5 ex]
	&	PI (\%)	&	80.0	&	80.7	&	90.5	&	90.3	&	84.2	&	84.2	&	83.0	&	84.8	\\ [0.5 ex]
	&	Rank	&	8	&	7	&	1	&	2	&	5	&	4	&	6	&	3	\\ [0.5 ex]
\midrule \multirow{4}{*}{Arrhythmia}	&	{Mean}	&	0.3483	&	0.3127	&	0.3004	&	0.2965	&	0.3348	&	0.3347	&	0.3295	&	0.2957	\\ [0.5 ex]
	&	{SD}	&	5.03E-03	&	3.22E-03	&	1.36E-02	&	1.34E-02	&	1.17E-02	&	1.15E-02	&	4.78E-03	&	9.69E-03	\\ [0.5 ex]
	&	PI (\%)	&	10.1	&	19.3	&	22.5	&	23.5	&	13.6	&	13.7	&	15.0	&	23.7	\\ [0.5 ex]
	&	Rank	&	8	&	4	&	3	&	2	&	7	&	6	&	5	&	1	\\ [0.5 ex]
\midrule \multirow{4}{*}{LSVT}	&	{Mean}	&	0.3350	&	0.2981	&	0.3240	&	0.3291	&	0.3286	&	0.3272	&	0.2919	&	0.1383	\\ [0.5 ex]
	&	{SD}	&	8.46E-03	&	1.85E-02	&	1.48E-02	&	8.09E-03	&	1.67E-02	&	1.68E-02	&	2.73E-02	&	1.60E-02	\\ [0.5 ex]
	&	PI (\%)	&	\textit{-70.5}	&	\textit{-51.7}	&	\textit{-64.9}	&	\textit{-67.4}	&	\textit{-67.2}	&	\textit{-66.5}	&	\textit{-48.6}	&	29.6	\\ [0.5 ex]
	&	Rank	&	8	&	3	&	4	&	7	&	6	&	5	&	2	&	1	\\ [0.5 ex]
\midrule \multirow{4}{*}{SECOM}	&	{Mean}	&	0.0652	&	0.0650	&	0.0643	&	0.0642	&	0.0649	&	0.0648	&	0.0645	&	0.0638	\\ [0.5 ex]
	&	{SD}	&	4.23E-04	&	3.58E-04	&	4.74E-04	&	4.46E-04	&	4.87E-04	&	5.75E-04	&	3.93E-04	&	7.96E-04	\\ [0.5 ex]
	&	PI (\%)	&	5.4	&	5.7	&	6.7	&	6.8	&	5.9	&	5.9	&	6.3	&	7.4	\\ [0.5 ex]
	&	Rank	&	8	&	7	&	3	&	2	&	6	&	5	&	4	&	1	\\ [0.5 ex]
\midrule \multicolumn{2}{c}{\textbf{Average Rank}}	&			6.8	&	4.1	&	2.8	&	3.3	&	6.6	&	6.0	&	3.3	&	2.1	\\ [0.5 ex]
\multicolumn{2}{c}{\textbf{Overall Rank}}	&			8	&	5	&	2	&	3	&	7	&	6	&	4	&	1	\\ [0.5 ex]

\bottomrule
\end{tabular}
 \begin{tablenotes}
      \tiny
      \item `Mean'and 'SD' - Mean and standard deviation of classification error over 40 runs
      \item `PI'- Percentage improvement in the classification error relative to the classification error achieved with all features
    \end{tablenotes}
  \end{threeparttable}
\end{adjustbox}
\end{table}

Further, \emph{paired t-tests} were used to evaluate the \emph{statistical significance} of the results relative to the compared algorithm. The tests were evaluated for $95\%$ \textit{confidence interval} ($p=0.05$) between the results obtained by 2D-UPSO and the compared algorithm. The results of this test are given in Table \ref{t:t} (for both NB and k-NN) wherein the entries with `+' indicates that the performance of 2D-UPSO is \textit{significantly better} than the compared algorithm while entries with `$-$' indicates \textit{otherwise}. For some datasets, the compared algorithms have similar results which are shown in Table \ref{t:t} by `=' entries. For several datasets where the test results are inconclusive, they are indicated by `?' in Table \ref{t:t}.  

The results of \textit{paired t-test} indicate that, irrespective of the induction algorithms, the results obtained with 2D-UPSO are \emph{significantly better} than GA, ErFS and PSO-42 for all the datasets investigated in the present study. Moreover, with NB classifier, for most of the datasets, the 2D-UPSO significantly outperformed ACO (12 out 16), BPSO (9 out of 16), CBPSO (11 out of 16) and chBPSO (11 out of 16), as shown in Table \ref{t:t}. The performance of the 2D-UPSO is similar with k-NN classifier as seen in (Table \ref{t:t}); with k-NN based classifier UPSO outperformed BPSO and CBPSO on eight datasets, chBPSO on nine datasets and ACO on ten datasets. 
\begin{table}[t]
\centering
\scriptsize
\begin{adjustbox}{max width=0.7\textwidth}
\begin{threeparttable}
\caption{Paired t-test of 2D-UPSO with the compared algorithms for 95\% confidence interval}
\label{t:t} 
\centering
\small
\begin{tabular}{c c c c c c c c c c c c c c c} 
 \toprule 
 \multirow{2}{*}{\textbf{Dataset}} & \multicolumn{2}{c}{\textbf{Vs. GA}} & \multicolumn{2}{c}{\textbf{Vs. ACO}} & \multicolumn{2}{c}{\textbf{Vs. BPSO}} & \multicolumn{2}{c}{\textbf{Vs. CBPSO}} & \multicolumn{2}{c}{\textbf{Vs. ErFS}} & \multicolumn{2}{c}{\textbf{Vs. PSO-42}} & \multicolumn{2}{c}{\textbf{Vs. chBPSO}}\\
& {NB} & {k-NN} & {NB} & {k-NN} & {NB} & {k-NN} & {NB} & {k-NN} & {NB} & {k-NN} & {NB} & {k-NN} & {NB} & {k-NN}\\ 
 \toprule
   
Wine	&	+	&	+	&	-	&	-	&	?	&	+	&	+	&	+	&	+	&	+	&	+	&	+	&	-	&	+	\\
Australian	&	+	&	+	&	-	&	-	&	+	&	+	&	+	&	+	&	+	&	+	&	+	&	+	&	?	&	+	\\
Zoo	&	+	&	=	&	=	&	=	&	=	&	=	&	=	&	=	&	+	&	+	&	+	&	+	&	=	&	=	\\
Vehicle	&	+	&	+	&	+	&	-	&	+	&	+	&	?	&	+	&	+	&	+	&	+	&	+	&	?	&	-	\\
German	&	+	&	+	&	-	&	?	&	-	&	?	&	?	&	?	&	+	&	+	&	+	&	+	&	-	&	+	\\
WDBC	&	+	&	+	&	+	&	-	&	+	&	+	&	+	&	+	&	+	&	+	&	+	&	+	&	+	&	-	\\
Ionosphere	&	+	&	+	&	+	&	+	&	+	&	+	&	+	&	+	&	+	&	+	&	+	&	+	&	+	&	+	\\
Lung Cancer	&	+	&	+	&	+	&	?	&	-	&	-	&	?	&	-	&	+	&	+	&	+	&	+	&	+	&	+	\\
Sonar	&	+	&	+	&	+	&	+	&	-	&	-	&	?	&	-	&	+	&	+	&	+	&	+	&	+	&	+	\\
Ozone	&	+	&	+	&	+	&	+	&	?	&	+	&	+	&	+	&	+	&	+	&	+	&	+	&	+	&	+	\\
Libras	&	+	&	+	&	+	&	+	&	-	&	-	&	-	&	-	&	+	&	+	&	+	&	+	&	+	&	+	\\
Hill	&	+	&	+	&	-	&	+	&	+	&	-	&	+	&	-	&	+	&	+	&	+	&	+	&	-	&	+	\\
Musk-1	&	+	&	+	&	+	&	+	&	+	&	-	&	+	&	-	&	+	&	+	&	+	&	+	&	+	&	+	\\
Arrhythmia 	&	+	&	+	&	+	&	+	&	+	&	?	&	+	&	?	&	+	&	+	&	+	&	+	&	+	&	+	\\
LSVT	&	+	&	+	&	+	&	+	&	+	&	+	&	+	&	+	&	+	&	+	&	+	&	+	&	+	&	+	\\
SECOM	&	+	&	+	&	+	&	+	&	?	&	+	&	+	&	+	&	+	&	+	&	+	&	+	&	+	&	+	\\
   
 \bottomrule
\end{tabular}
\begin{tablenotes}
      \scriptsize
      \item `$+ / -$'   :   UPSO results are significantly better/worse than the compared algorithm,
      \item `=' :   results of the compared algorithms are equal, `?' :   indeterminate result
    \end{tablenotes}
  \end{threeparttable}
\end{adjustbox}
\end{table}
\begin{table}
\small
\centering
\scalebox{0.85}{
\begin{threeparttable}
\caption{Average feature subset cardinality ($\xi_{avg}$) over 40 runs (NB classifier)}
\label{t:l:nb} 
\centering
\begin{tabular}{c c c c c c c c c c} 
 \toprule
 \multirow{1}{*}{\textbf{Dataset}} & \multirow{1}{*}{\textbf{Results}} & {\textbf{GA}} & {\textbf{ACO}} & {\textbf{BPSO}} & {\textbf{CBPSO}} & {\textbf{ErFS}} & {\textbf{PSO-42}} & {\textbf{chBPSO}} & {\textbf{2D-UPSO}} \\ [1.0ex] 
 \toprule
    
  \multirow{3}{*}{Wine}	&	{$\xi_{avg}$}	&	6.70	&	4.00	&	4.23	&	4.38	&	5.10	&	5.23	&	4.00	&	4.15	\\ [0.2ex]
	&	$\Xi \ (\%)$	&	48.5	&	69.2	&	67.5	&	66.3	&	60.8	&	59.8	&	69.2	&	68.1	\\ [0.2ex]
	&	Rank	&	8	&	1	&	4	&	5	&	6	&	7	&	1	&	3	\\ [0.2ex]
    \midrule \multirow{3}{*}{Australian}	&	{$\xi_{avg}$}	&	7.20	&	11.00	&	10.25	&	10.40	&	9.95	&	10.45	&	10.80	&	10.90	\\ [0.2ex]
	&	$\Xi \ (\%)$	&	48.6	&	21.4	&	26.8	&	25.7	&	28.9	&	25.4	&	22.9	&	22.1	\\ [0.2ex]
	&	Rank	&	1	&	8	&	3	&	4	&	2	&	5	&	6	&	7	\\ [0.2ex]
    \midrule \multirow{3}{*}{Zoo}	&	{$\xi_{avg}$}	&	8.03	&	9.10	&	10.15	&	9.85	&	11.68	&	10.85	&	9.23	&	9.58	\\ [0.2ex]
	&	$\Xi \ (\%)$	&	52.8	&	46.5	&	40.3	&	42.1	&	31.3	&	36.2	&	45.7	&	43.7	\\ [0.2ex]
	&	Rank	&	1	&	2	&	6	&	5	&	8	&	7	&	3	&	4	\\ [0.2ex]
    \midrule \multirow{3}{*}{Vehicle} 	&	{$\xi_{avg}$}	&	8.48	&	8.00	&	8.08	&	8.13	&	9.08	&	9.28	&	7.90	&	8.03	\\ [0.2ex]
	&	$\Xi \ (\%)$	&	52.9	&	55.6	&	55.1	&	54.9	&	49.6	&	48.5	&	56.1	&	55.4	\\ [0.2ex]
	&	Rank	&	6	&	2	&	4	&	5	&	7	&	8	&	1	&	3	\\ [0.2ex]
    \midrule \multirow{3}{*}{German}	&	{$\xi_{avg}$}	&	12.25	&	11.35	&	11.10	&	11.23	&	12.58	&	12.68	&	11.33	&	11.05	\\ [0.2ex]
	&	$\Xi \ (\%)$	&	49.0	&	52.7	&	53.8	&	53.2	&	47.6	&	47.2	&	52.8	&	54.0	\\ [0.2ex]
	&	Rank	&	6	&	5	&	2	&	3	&	7	&	8	&	4	&	1	\\ [0.2ex]
    \midrule \multirow{3}{*}{WDBC}	&	{$\xi_{avg}$}	&	14.78	&	8.68	&	13.60	&	18.60	&	25.50	&	7.90	&	6.50	&	3.80	\\ [0.2ex]
	&	$\Xi \ (\%)$	&	50.8	&	71.1	&	54.7	&	38.0	&	15.0	&	73.7	&	78.3	&	87.3	\\ [0.2ex]
	&	Rank	&	6	&	4	&	5	&	7	&	8	&	3	&	2	&	1	\\ [0.2ex]
    \midrule \multirow{3}{*}{Ionosphere}	&	{$\xi_{avg}$}	&	17.83	&	3.43	&	3.18	&	3.40	&	3.48	&	4.03	&	3.65	&	1.63	\\ [0.2ex]
	&	$\Xi \ (\%)$	&	47.6	&	89.9	&	90.7	&	90.0	&	89.8	&	88.2	&	89.3	&	95.2	\\ [0.2ex]
	&	Rank	&	8	&	4	&	2	&	3	&	5	&	7	&	6	&	1	\\ [0.2ex]
    \midrule \multirow{3}{*}{Lung Cancer}	&	{$\xi_{avg}$}	&	28.75	&	23.50	&	24.53	&	24.60	&	28.50	&	26.73	&	23.80	&	18.28	\\ [0.2ex]
	&	$\Xi \ (\%)$	&	50.4	&	59.5	&	57.7	&	57.6	&	50.9	&	53.9	&	59.0	&	68.5	\\ [0.2ex]
	&	Rank	&	8	&	2	&	4	&	5	&	7	&	6	&	3	&	1	\\ [0.2ex]
    \midrule \multirow{3}{*}{Sonar}	&	{$\xi_{avg}$}	&	30.30	&	22.03	&	15.45	&	18.38	&	20.20	&	15.48	&	18.48	&	6.53	\\ [0.2ex]
	&	$\Xi \ (\%)$	&	49.5	&	63.3	&	74.3	&	69.4	&	66.3	&	74.2	&	69.2	&	89.1	\\ [0.2ex]
	&	Rank	&	8	&	7	&	2	&	4	&	6	&	3	&	5	&	1	\\ [0.2ex]
    \midrule \multirow{3}{*}{Ozone}	&	{$\xi_{avg}$}	&	36.58	&	28.58	&	29.98	&	30.13	&	27.53	&	24.20	&	27.85	&	16.50	\\ [0.2ex]
	&	$\Xi \ (\%)$	&	49.2	&	60.3	&	58.4	&	58.2	&	61.8	&	66.4	&	61.3	&	77.1	\\ [0.2ex]
	&	Rank	&	8	&	5	&	6	&	7	&	3	&	2	&	4	&	1	\\ [0.2ex]
    \midrule \multirow{3}{*}{Libras}	&	{$\xi_{avg}$}	&	45.73	&	40.55	&	37.33	&	37.38	&	45.45	&	44.18	&	38.05	&	28.53	\\ [0.2ex]
	&	$\Xi \ (\%)$	&	50.3	&	55.9	&	59.4	&	59.4	&	50.6	&	52.0	&	58.6	&	69.0	\\ [0.2ex]
	&	Rank	&	8	&	5	&	2	&	3	&	7	&	6	&	4	&	1	\\ [0.2ex]
    \midrule \multirow{3}{*}{Hill}	&	{$\xi_{avg}$}	&	50.68	&	44.68	&	50.60	&	51.75	&	50.80	&	45.10	&	45.33	&	43.78	\\ [0.2ex]
	&	$\Xi \ (\%)$	&	49.3	&	55.3	&	49.4	&	48.3	&	49.2	&	54.9	&	54.7	&	56.2	\\ [0.5ex]
	&	Rank	&	6	&	2	&	5	&	8	&	7	&	3	&	4	&	1	\\ [0.2ex]
    \midrule \multirow{3}{*}{Musk-1}	&	{$\xi_{avg}$}	&	83.65	&	78.15	&	78.95	&	79.60	&	86.08	&	90.48	&	79.08	&	23.48	\\ [0.5ex]
	&	$\Xi \ (\%)$	&	49.6	&	52.9	&	52.4	&	52.0	&	48.1	&	45.5	&	52.4	&	85.9	\\ [0.5ex]
	&	Rank	&	6	&	2	&	3	&	5	&	7	&	8	&	4	&	1	\\ [0.2ex]
    \midrule \multirow{3}{*}{Arrhythmia}	&	{$\xi_{avg}$}	&	138.13	&	135.70	&	135.48	&	134.15	&	157.10	&	155.32	&	134.25	&	6.90	\\ [0.2ex]
	&	$\Xi \ (\%)$	&	50.5	&	51.4	&	51.4	&	51.9	&	43.7	&	44.3	&	51.9	&	97.5	\\ [0.2ex]
	&	Rank	&	6	&	5	&	4	&	2	&	8	&	7	&	3	&	1	\\ [0.2ex]
    \midrule \multirow{3}{*}{LSVT}	&	{$\xi_{avg}$}	&	157.50	&	151.25	&	154.63	&	157.40	&	191.60	&	179.18	&	147.43	&	62.95	\\ [0.2ex]
	&	$\Xi \ (\%)$	&	49.2	&	51.2	&	50.1	&	49.2	&	38.2	&	42.2	&	52.4	&	79.7	\\ [0.2ex]
	&	Rank	&	6	&	3	&	4	&	5	&	8	&	7	&	2	&	1	\\ [0.2ex]
    \midrule \multirow{3}{*}{SECOM}	&	{$\xi_{avg}$}	&	294.52	&	293.20	&	312.40	&	307.80	&	343.73	&	330.13	&	292.83	&	203.61	\\ [0.2ex]
	&	$\Xi \ (\%)$	&	50.2	&	50.4	&	47.1	&	47.9	&	41.8	&	44.1	&	50.5	&	65.5	\\ [0.2ex]
	&	Rank	&	4	&	3	&	6	&	5	&	8	&	7	&	2	&	1	\\ [0.2ex]
    \midrule\multicolumn{2}{c}{\textbf{Average Rank}}			&	6.0	&	3.8	&	3.9	&	4.8	&	6.5	&	5.9	&	3.4	&	1.8	\\ [0.2ex]
    \multicolumn{2}{c}{\textbf{Final Rank}}			&	7	&	3	&	4	&	5	&	8	&	6	&	2	&	1	\\ [0.2ex]
    
 \bottomrule
\end{tabular}
 \begin{tablenotes}
      \tiny
      \item `$\xi_{avg}$' - Average length of the feature subset over 40 runs
    \end{tablenotes}
  \end{threeparttable}
}
\end{table}
\subsection{Reduction in Subset Cardinality}

The average subset cardinality obtained by all the compared algorithms are shown in Table \ref{t:l:nb} for NB classifier and in Table \ref{t:l:knn} for k-NN classifier. For each dataset, the \textit{average cardinality} ($\xi_{avg}$) of the subsets for each algorithm, after 40 independent runs, is shown in Tables \ref{t:l:nb} and \ref{t:l:knn}. Further, the \textit{percentage reduction} in the cardinality with respect to total number of features ($n$) is used to \textit{rank} each algorithm. This metric is denoted here as $\Xi \ (\%)$ and is given by
\begin{equation*}
    \Xi (\%) = \frac {n - \xi_{avg}}{n} \times 100
\end{equation*}

Since the cardinality is not explicitly included in the criterion function, $J(\cdotp)$, (as seen in (\ref{eq:fitness})), any reduction in the cardinality is the direct consequence of the algorithm's ability to distinguish irrelevant features. The results of the comparative evaluation confirms the utility of the cardinality information; 2D-UPSO was able find \textit{smallest} feature subset (as seen in Tables \ref{t:l:nb} and \ref{t:l:knn}) with a \textit{better} classification performance (Table \ref{t:er:nb}-\ref{t:er:knn}) for most of the datasets. Out of 16 benchmark datasets, 2D-UPSO obtained smallest subset for twelve datasets with NB classifier (Table \ref{t:l:nb}) and for ten datasets with k-NN classifier (Table \ref{t:l:knn}) and achieved overall best performance with both the classifiers. Among compared algorithms, ACO and chBPSO performed better with both the classifiers. Further, the results indicate that GA is more susceptible to the change in the search landscape caused by the induction algorithm.This is evident by the change in GA's relative performance, from the seventh rank with NB classifier (Table \ref{t:l:nb}) to the fourth rank with k-NN classifier (Table \ref{t:l:knn}). 
\begin{table}
\small
\centering
\scalebox{0.85}{
\begin{threeparttable}
\caption{Average feature subset cardinality ($\xi_{avg}$) over 40 runs (k-NN classifier)}
\label{t:l:knn} 
\centering
\begin{tabular}{c c c c c c c c c c} 
 \toprule
 \multirow{1}{*}{\textbf{Dataset}} & \multirow{1}{*}{\textbf{Results}} & {\textbf{GA}} & {\textbf{ACO}} & {\textbf{BPSO}} & {\textbf{CBPSO}} & {\textbf{ErFS}} & {\textbf{PSO-42}} & {\textbf{chBPSO}} & {\textbf{2D-UPSO}} \\ [1.0ex] 
 \toprule
   
   \multirow{3}{*}{Wine}	&	{$\xi_{avg}$}	&	7.05	&	6.00	&	8.60	&	8.40	&	10.05	&	9.38	&	6.43	&	7.50	\\ [0.2ex]
	&	$\Xi \ (\%)$	&	45.8	&	53.8	&	33.8	&	35.4	&	22.7	&	27.9	&	50.6	&	42.3	\\ [0.2ex]
	&	Rank	&	3	&	1	&	6	&	5	&	8	&	7	&	2	&	4	\\ [0.2ex]
    \midrule \multirow{3}{*}{Australian}	&	{$\xi_{avg}$}	&	6.55	&	2.68	&	4.98	&	4.33	&	5.43	&	5.55	&	2.68	&	2.20	\\ [0.2ex]
	&	$\Xi \ (\%)$	&	53.2	&	80.9	&	64.5	&	69.1	&	61.3	&	60.4	&	80.9	&	84.3	\\ [0.2ex]
	&	Rank	&	8	&	2	&	5	&	4	&	6	&	7	&	2	&	1	\\ [0.2ex]
    \midrule \multirow{3}{*}{Zoo}	&	{$\xi_{avg}$}	&	8.08	&	7.78	&	8.73	&	8.35	&	9.33	&	7.23	&	7.48	&	4.18	\\ [0.2ex]
	&	$\Xi \ (\%)$	&	52.5	&	54.3	&	48.7	&	50.9	&	45.1	&	57.5	&	56.0	&	75.4	\\ [0.2ex]
	&	Rank	&	5	&	4	&	7	&	6	&	8	&	2	&	3	&	1	\\ [0.2ex]
    \midrule \multirow{3}{*}{Vehicle} 	&	{$\xi_{avg}$}	&	8.85	&	8.33	&	9.60	&	9.48	&	12.15	&	11.38	&	8.70	&	8.83	\\ [0.2ex]
	&	$\Xi \ (\%)$	&	50.8	&	53.8	&	46.7	&	47.4	&	32.5	&	36.8	&	51.7	&	51.0	\\ [0.2ex]
	&	Rank	&	4	&	1	&	6	&	5	&	8	&	7	&	2	&	3	\\ [0.2ex]
    \midrule \multirow{3}{*}{German}	&	{$\xi_{avg}$}	&	12.50	&	10.78	&	12.20	&	11.83	&	14.60	&	15.05	&	10.85	&	11.75	\\ [0.2ex]
	&	$\Xi \ (\%)$	&	47.9	&	55.1	&	49.2	&	50.7	&	39.2	&	37.3	&	54.8	&	51.0	\\ [0.2ex]
	&	Rank	&	6	&	1	&	5	&	4	&	7	&	8	&	2	&	3	\\ [0.2ex]
    \midrule \multirow{3}{*}{WDBC}	&	{$\xi_{avg}$}	&	14.93	&	14.18	&	15.23	&	15.38	&	17.38	&	15.13	&	14.83	&	13.10	\\ [0.2ex]
	&	$\Xi \ (\%)$	&	50.3	&	52.8	&	49.3	&	48.8	&	42.1	&	49.6	&	50.6	&	56.3	\\ [0.2ex]
	&	Rank	&	4	&	2	&	6	&	7	&	8	&	5	&	3	&	1	\\ [0.2ex]
    \midrule \multirow{3}{*}{Ionosphere}	&	{$\xi_{avg}$}	&	16.80	&	13.73	&	15.45	&	14.25	&	16.53	&	16.55	&	12.95	&	5.25	\\ [0.2ex]
	&	$\Xi \ (\%)$	&	50.6	&	59.6	&	54.6	&	58.1	&	51.4	&	51.3	&	61.9	&	84.6	\\ [0.2ex]
	&	Rank	&	8	&	3	&	5	&	4	&	6	&	7	&	2	&	1	\\ [0.2ex]
    \midrule \multirow{3}{*}{Lung Cancer}	&	{$\xi_{avg}$}	&	28.40	&	28.60	&	31.63	&	31.75	&	36.68	&	34.43	&	28.65	&	33.38	\\ [0.2ex]
	&	$\Xi \ (\%)$	&	51.0	&	50.7	&	45.5	&	45.3	&	36.8	&	40.6	&	50.6	&	42.5	\\ [0.2ex]
	&	Rank	&	1	&	2	&	4	&	5	&	8	&	7	&	3	&	6	\\ [0.2ex]
    \midrule \multirow{3}{*}{Sonar}	&	{$\xi_{avg}$}	&	30.10	&	28.60	&	30.88	&	31.03	&	34.70	&	32.63	&	27.83	&	28.50	\\ [0.2ex]
	&	$\Xi \ (\%)$	&	49.8	&	52.3	&	48.5	&	48.3	&	42.2	&	45.6	&	53.6	&	52.5	\\ [0.2ex]
	&	Rank	&	4	&	3	&	5	&	6	&	8	&	7	&	1	&	2	\\ [0.2ex]
    \midrule \multirow{3}{*}{Ozone}	&	{$\xi_{avg}$}	&	35.98	&	32.63	&	37.30	&	38.45	&	37.23	&	36.58	&	31.95	&	23.30	\\ [0.2ex]
	&	$\Xi \ (\%)$	&	50.0	&	54.7	&	48.2	&	46.6	&	48.3	&	49.2	&	55.6	&	67.6	\\ [0.2ex]
	&	Rank	&	4	&	3	&	7	&	8	&	6	&	5	&	2	&	1	\\ [0.2ex]
    \midrule \multirow{3}{*}{Libras}	&	{$\xi_{avg}$}	&	43.95	&	41.38	&	43.33	&	44.13	&	52.43	&	48.95	&	41.63	&	34.05	\\ [0.2ex]
	&	$\Xi \ (\%)$	&	52.2	&	55.0	&	52.9	&	52.0	&	43.0	&	46.8	&	54.8	&	63.0	\\ [0.2ex]
	&	Rank	&	5	&	2	&	4	&	6	&	8	&	7	&	3	&	1	\\ [0.2ex]
    \midrule \multirow{3}{*}{Hill}	&	{$\xi_{avg}$}	&	48.95	&	42.55	&	45.50	&	46.53	&	47.78	&	47.15	&	45.35	&	16.23	\\ [0.5ex]
	&	$\Xi \ (\%)$	&	51.1	&	57.5	&	54.5	&	53.5	&	52.2	&	52.9	&	54.7	&	83.8	\\ [0.2ex]
	&	Rank	&	8	&	2	&	4	&	5	&	7	&	6	&	3	&	1	\\ [0.2ex]
    \midrule \multirow{3}{*}{Musk-1}	&	{$\xi_{avg}$}	&	84.05	&	84.83	&	93.10	&	95.05	&	116.13	&	115.80	&	85.75	&	113.20	\\ [0.2ex]
	&	$\Xi \ (\%)$	&	49.4	&	48.9	&	43.9	&	42.7	&	30.0	&	30.2	&	48.3	&	31.8	\\ [0.2ex]
	&	Rank	&	1	&	2	&	4	&	5	&	8	&	7	&	3	&	6	\\ [0.2ex]
    \midrule \multirow{3}{*}{Arrhythmia}	&	{$\xi_{avg}$}	&	142.20	&	132.82	&	144.58	&	138.80	&	166.82	&	163.20	&	136.72	&	62.80	\\ [0.2ex]
	&	$\Xi \ (\%)$	&	49.0	&	52.4	&	48.2	&	50.3	&	40.2	&	41.5	&	51.0	&	77.5	\\ [0.2ex]
	&	Rank	&	5	&	2	&	6	&	4	&	8	&	7	&	3	&	1	\\ [0.5ex]
    \midrule \multirow{3}{*}{LSVT}	&	{$\xi_{avg}$}	&	152.83	&	151.88	&	164.53	&	166.15	&	175.45	&	132.23	&	151.88	&	16.50	\\ [0.2ex]
	&	$\Xi \ (\%)$	&	50.7	&	51.0	&	46.9	&	46.4	&	43.4	&	57.3	&	51.0	&	94.7	\\ [0.2ex]
	&	Rank	&	5	&	4	&	6	&	7	&	8	&	2	&	3	&	1	\\ [0.5ex]
    \midrule \multirow{3}{*}{SECOM}	&	{$\xi_{avg}$}	&	294.05	&	297.68	&	325.83	&	331.77	&	372.23	&	356.77	&	292.38	&	224.88	\\ [0.2ex]
	&	$\Xi \ (\%)$	&	50.2	&	49.6	&	44.9	&	43.9	&	37.0	&	39.6	&	50.5	&	62.0	\\ [0.2ex]
	&	Rank	&	3	&	4	&	5	&	6	&	8	&	7	&	2	&	1	\\ [0.5ex]
    \midrule\multicolumn{2}{c}{\textbf{Average Rank}}			&	4.63	&	2.38	&	5.31	&	5.44	&	7.50	&	6.13	&	2.44	&	2.13	\\ [0.2ex]
    \multicolumn{2}{c}{\textbf{Final Rank}}			&	4	&	2	&	5	&	6	&	8	&	7	&	3	&	1	\\ [0.2ex]
   
 \bottomrule
\end{tabular}
 \begin{tablenotes}
      \tiny
      \item `$\xi_{avg}$' - Average length of the feature subset over 40 runs
    \end{tablenotes}
  \end{threeparttable}
}
\end{table}

\subsection{Timing Analysis}

The time complexity of the algorithms is usually expressed in terms of `\textit{Big-O-notations}'. However, such analysis will provide a very loose estimate of the \textit{`time-complexity'}, especially for \textit{meta-heuristic wrappers} \cite{Oh:Lee:2004}. This is because, in the wrapper approach, for a given pair of dataset and algorithm, the `\textit{subset evaluation time}' is expected be a major part of the total run time. Further, the subset evaluation time is dependent on the cardinality of the selected subsets. The algorithm, which selects smaller subsets throughout the run, will have a lower subset evaluation time and, therefore, a lower total run time. Since all the compared algorithms, alongwith 2D-UPSO, are in essence \textit{meta-heuristic wrappers}, in this study, a different approach based on the procedure outlined in \cite{Suganthan:Hansen:2005}, is used to investigate the time-complexity of the algorithms. 

For a given dataset, \textit{ten independent runs} of each of the compared algorithm were carried out. Each run was set to execute for 6000 Function Evaluations (FEs). During each run, two time measurements were recorded: `\textit{total run time}' of the algorithm ($t$) and the time spent in `\textit{subset evaluation}' ($t_2$). From these two time measurements, the time spent only on the algorithm ($t_1$) was estimated as $t_1=t-t_2$. The results of this test are shown in Table~\ref{t:time:nb} (with NB classifier) and Table~\ref{t:time:knn} (with k-NN classifier), wherein `$T_1$', `$T_2$' and `$T$' represent the average values of `\textit{algorithm run time}' ($t_1$), `\textit{subset evaluation time}' ($t_2$) and `\textit{total run time}' ($t$) obtained over ten independent runs. For this test, GA, ACO and BPSO were selected for the benchmarking; since the remaining PSO variants are based on BPSO, their performance in terms of time is not likely to significantly differ from BPSO. Further, each algorithm was ranked based on the corresponding average total run time ($T$), \textit{i.e.}, for a given dataset, algorithm with the lowest $T$ is ranked `1'.  

The results of the timing analysis indicate that the `\textit{subset evaluation time}' ($T_2$) is indeed the major contributor to total run time and it is significantly higher than `\textit{algorithm run time}' ($T_1$), \textit{i.e.}, $T_2 \gg T_1$. The comparative analysis indicates that for majority of the datasets, the  lowest subset evaluation time ($T_2$) was obtained with 2D-UPSO (on 11 datasets with NB and 9 datasets with k-NN). It is obvious from the results (Table~\ref{t:time:nb} and~\ref{t:time:knn}), that the subset evaluation time ($T_2$) in the the proposed 2D-UPSO is significantly less. This can be ascribed to efficient use of cardinality information. This observation is further corroborated by the subset length results shown in Tables~\ref{t:l:nb} and~\ref{t:l:knn}. Since $T_2$ dominates the total run time, $T$, for most of the datasets, 2D-UPSO is faster than all the compared algorithms and provides overall best performance with both NB and k-NN classifiers (Table~\ref{t:time:nb} and~\ref{t:time:knn}). 

Note that the `\textit{algorithm run time}' ($T_1$) can be used as a metric to measure the time complexity of the algorithm. The comparative evaluation of $T_1$ (Table~\ref{t:time:nb} and~\ref{t:time:knn}) indicates that, compared to BPSO, the time-complexity of 2D-UPSO does not increase significantly despite the additional computations associated with the integration of cardinality information. For example, even for a larger dataset such as `SECOM', there is aproximately one second increase in algorithm run time of 2D-UPSO compared to BPSO, as seen in Tables~\ref{t:time:nb} and~\ref{t:time:knn}. Given that the total run time for this dataset is in the range of thousands of seconds (approx. 2000 s witn NB, Table~\ref{t:time:nb}; approx. 5000 s with k-NN, Table~\ref{t:time:knn}), one second increase in algorithm run time is almost negligible. Finally, on the basis of $T_1$, time complexity of the compared algorithms can be ranked as follows: ACO$<$GA$<$BPSO$<$2D-UPSO. Nevertheless, due to consistent discovery of smaller subsets and the fact that $T_2 \gg T_1$, 2D-UPSO is able to complete the search in shorter time compared to other algorithms.

\begin{table}
    \caption*{}
    \begin{minipage}{.5\linewidth}
    
    \scriptsize
    \centering
    \scalebox{0.75}{
    \begin{threeparttable}
    \caption{Average computational times (in seconds) with NB}
    \label{t:time:nb} 
    \centering
    \begin{tabular}{c c c c c c} 
    \toprule
    \multirow{1}{*}{\textbf{Dataset}} & \multirow{1}{*}{\textbf{Results}} & {\textbf{GA}} & {\textbf{ACO}} & {\textbf{BPSO}} & {\textbf{2D-UPSO}} \\ [1.0ex] 
    
   \midrule \multirow{4}{*}{Wine}	&	${T_1}$	&	0.14	&	0.09	&	0.19	&	0.29	\\ [0.1 ex]
	&	${T_2}$	&	108.05	&	107.05	&	129.56	&	105.82	\\ [0.1 ex]
	&	${T}$	&	108.19	&	107.15	&	129.75	&	106.11	\\ [0.1 ex]
	&	Rank	&	3	&	2	&	4	&	1	\\ [0.1 ex]
    \midrule \multirow{4}{*}{Australian}	&	${T_1}$	&	0.10	&	0.09	&	0.25	&	0.37	\\ [0.1 ex]
	&	${T_2}$	&	337.02	&	341.94	&	338.63	&	334.39	\\ [0.1 ex]
	&	${T}$	&	337.12	&	342.03	&	338.88	&	334.76	\\ [0.1 ex]
	&	Rank	&	2	&	4	&	3	&	1	\\ [0.1 ex]
    \midrule \multirow{4}{*}{Zoo}	&	${T_1}$	&	0.14	&	0.06	&	0.21	&	0.23	\\ [0.1 ex]
	&	${T_2}$	&	89.78	&	88.42	&	88.05	&	87.43	\\ [0.1 ex]
	&	${T}$	&	89.92	&	88.47	&	88.26	&	87.66	\\ [0.1 ex]
	&	Rank	&	4	&	3	&	2	&	1	\\ [0.1 ex]
    \midrule \multirow{4}{*}{Vehicle}	&	${T_1}$	&	0.20	&	0.12	&	0.29	&	0.35	\\ [0.1 ex]
	&	${T_2}$	&	426.54	&	420.85	&	409.26	&	398.04	\\ [0.1 ex]
	&	${T}$	&	426.74	&	420.97	&	409.55	&	398.39	\\ [0.1 ex]
	&	Rank	&	4	&	3	&	2	&	1	\\ [0.1 ex]
    \midrule \multirow{4}{*}{German}	&	${T_1}$	&	0.16	&	0.13	&	0.29	&	0.37	\\ [0.1 ex]
	&	${T_2}$	&	462.05	&	454.38	&	445.60	&	432.21	\\ [0.1 ex]
	&	${T}$	&	462.22	&	454.51	&	445.89	&	432.58	\\ [0.1 ex]
	&	Rank	&	4	&	3	&	2	&	1	\\ [0.1 ex]
    \midrule \multirow{4}{*}{WDBC}	&	${T_1}$	&	0.20	&	0.13	&	0.24	&	0.36	\\ [0.1 ex]
	&	${T_2}$	&	280.66	&	272.82	&	268.01	&	261.40	\\ [0.1 ex]
	&	${T}$	&	280.86	&	272.94	&	268.24	&	261.76	\\ [0.1 ex]
	&	Rank	&	4	&	3	&	2	&	1	\\ [0.1 ex]
    \midrule \multirow{4}{*}{Ionosphere} 	&	${T_1}$	&	0.16	&	0.11	&	0.22	&	0.38	\\ [0.1 ex]
	&	${T_2}$	&	182.22	&	177.87	&	167.08	&	173.20	\\ [0.1 ex]
	&	${T}$	&	182.38	&	177.98	&	167.30	&	173.59	\\ [0.1 ex]
	&	Rank	&	4	&	3	&	1	&	2	\\ [0.1 ex]
    \midrule \multirow{4}{*}{Lung Cancer}	&	${T_1}$	&	0.17	&	0.08	&	0.17	&	0.28	\\ [0.1 ex]
	&	${T_2}$	&	41.41	&	40.86	&	40.64	&	40.87	\\ [0.1 ex]
	&	${T}$	&	41.58	&	40.93	&	40.82	&	41.15	\\ [0.1 ex]
	&	Rank	&	4	&	2	&	1	&	3	\\ [0.1 ex]
    \midrule \multirow{4}{*}{Sonar}	&	${T_1}$	&	0.18	&	0.07	&	0.22	&	0.36	\\ [0.1 ex]
	&	${T_2}$	&	126.49	&	119.18	&	112.39	&	115.03	\\ [0.1 ex]
	&	${T}$	&	126.67	&	119.24	&	112.61	&	115.39	\\ [0.1 ex]
	&	Rank	&	4	&	3	&	1	&	2	\\ [0.1 ex]
    \midrule \multirow{4}{*}{Ozone}	&	${T_1}$	&	0.19	&	0.10	&	0.37	&	0.47	\\ [0.1 ex]
	&	${T_2}$	&	936.76	&	886.96	&	829.57	&	815.79	\\ [0.1 ex]
	&	${T}$	&	936.95	&	887.06	&	829.94	&	816.26	\\ [0.1 ex]
	&	Rank	&	4	&	3	&	2	&	1	\\ [0.1 ex]
    \midrule \multirow{4}{*}{Libras}	&	${T_1}$	&	0.15	&	0.12	&	0.33	&	0.54	\\ [0.1 ex]
	&	${T_2}$	&	404.58	&	395.20	&	370.09	&	372.68	\\ [0.1 ex]
	&	${T}$	&	404.73	&	395.32	&	370.43	&	373.23	\\ [0.1 ex]
	&	Rank	&	4	&	3	&	1	&	2	\\ [0.1 ex]
    \midrule \multirow{4}{*}{Hill}	&	${T_1}$	&	0.18	&	0.16	&	0.37	&	0.53	\\ [0.1 ex]
	&	${T_2}$	&	332.20	&	321.75	&	318.54	&	321.13	\\ [0.1 ex]
	&	${T}$	&	332.38	&	321.91	&	318.91	&	321.66	\\ [0.1 ex]
	&	Rank	&	4	&	3	&	1	&	2	\\ [0.1 ex]
    \midrule \multirow{4}{*}{Musk-1}	&	${T_1}$	&	0.19	&	0.15	&	0.46	&	0.79	\\ [0.1 ex]
	&	${T_2}$	&	318.75	&	314.08	&	309.37	&	307.33	\\ [0.1 ex]
	&	${T}$	&	318.94	&	314.23	&	309.82	&	308.12	\\ [0.1 ex]
	&	Rank	&	4	&	3	&	2	&	1	\\ [0.1 ex]
    \midrule \multirow{4}{*}{Arrhythmia}	&	${T_1}$	&	0.26	&	0.16	&	0.53	&	0.98	\\ [0.1 ex]
	&	${T_2}$	&	749.30	&	724.52	&	681.77	&	605.49	\\ [0.1 ex]
	&	${T}$	&	749.56	&	724.68	&	682.30	&	606.47	\\ [0.1 ex]
	&	Rank	&	4	&	3	&	2	&	1	\\ [0.1 ex]
    \midrule \multirow{4}{*}{LSVT}	&	${T_1}$	&	0.27	&	0.21	&	0.45	&	1.06	\\ [0.1 ex]
	&	${T_2}$	&	128.80	&	130.07	&	130.66	&	125.06	\\ [0.1 ex]
	&	${T}$	&	129.08	&	130.27	&	131.11	&	126.13	\\ [0.1 ex]
	&	Rank	&	2	&	3	&	4	&	1	\\ [0.1 ex]
    \midrule \multirow{4}{*}{SECOM}	&	${T_1}$	&	0.29	&	0.29	&	0.70	&	1.79	\\ [0.1 ex]
	&	${T_2}$	&	2198.63	&	2049.13	&	2179.46	&	1962.57	\\ [0.1 ex]
	&	${T}$	&	2198.93	&	2049.43	&	2180.16	&	1964.36	\\ [0.1 ex]
	&	Rank	&	4	&	2	&	3	&	1	\\ [0.1 ex]
    \midrule \multicolumn{2}{c}{\textbf{Average Rank}}			&	3.69	&	2.88	&	2.06	&	1.38	\\ [0.1 ex]
    \multicolumn{2}{c}{\textbf{Overall Rank}}			&	4	&	3	&	2	&	1	\\ [0.1 ex]
    
    \bottomrule
    \end{tabular}
    \end{threeparttable}
    }  
    \end{minipage}%
    \begin{minipage}{.5\linewidth}
    \scriptsize
    \centering
    \scalebox{0.75}{
    \begin{threeparttable}
    \caption{Average computational times (in seconds) with k-NN}
    \label{t:time:knn} 
    \centering
    \begin{tabular}{c c c c c c} 
    \toprule
    \multirow{1}{*}{\textbf{Dataset}} & \multirow{1}{*}{\textbf{Results}} & {\textbf{GA}} & {\textbf{ACO}} & {\textbf{BPSO}} & {\textbf{2D-UPSO}} \\ [1.0ex] 
    
   \midrule \multirow{4}{*}{Wine}	&	${T_1}$	&	0.20	&	0.12	&	0.32	&	0.30	\\ [0.1 ex]
	&	${T_2}$	&	632.02	&	643.19	&	636.96	&	633.79	\\ [0.1 ex]
	&	${T}$	&	632.22	&	643.31	&	637.28	&	634.08	\\ [0.1 ex]
	&	Rank	&	1	&	4	&	3	&	2	\\ [0.1 ex]
    \midrule \multirow{4}{*}{Australian}	&	${T_1}$	&	0.11	&	0.09	&	0.32	&	0.34	\\ [0.1 ex]
	&	${T_2}$	&	711.98	&	711.71	&	705.15	&	690.67	\\ [0.1 ex]
	&	${T}$	&	712.10	&	711.81	&	705.47	&	691.01	\\ [0.1 ex]
	&	Rank	&	4	&	3	&	2	&	1	\\ [0.1 ex]
    \midrule \multirow{4}{*}{Zoo}	&	${T_1}$	&	0.17	&	0.12	&	0.35	&	0.38	\\ [0.1 ex]
	&	${T_2}$	&	669.84	&	677.25	&	668.89	&	670.55	\\ [0.1 ex]
	&	${T}$	&	670.01	&	677.37	&	669.24	&	670.93	\\ [0.1 ex]
	&	Rank	&	2	&	4	&	1	&	3	\\ [0.1 ex]
    \midrule \multirow{4}{*}{Vehicle}	&	${T_1}$	&	0.14	&	0.12	&	0.29	&	0.36	\\ [0.1 ex]
	&	${T_2}$	&	774.86	&	780.47	&	781.58	&	760.94	\\ [0.1 ex]
	&	${T}$	&	775.00	&	780.58	&	781.87	&	761.30	\\ [0.1 ex]
	&	Rank	&	2	&	3	&	4	&	1	\\ [0.1 ex]
    \midrule \multirow{4}{*}{German}	&	${T_1}$	&	0.21	&	0.11	&	0.38	&	0.31	\\ [0.1 ex]
	&	${T_2}$	&	837.72	&	837.23	&	841.53	&	820.75	\\ [0.1 ex]
	&	${T}$	&	837.92	&	837.34	&	841.90	&	821.06	\\ [0.1 ex]
	&	Rank	&	3	&	2	&	4	&	1	\\ [0.1 ex]
    \midrule \multirow{4}{*}{WDBC}	&	${T_1}$	&	0.17	&	0.10	&	0.29	&	0.40	\\ [0.1 ex]
	&	${T_2}$	&	743.63	&	744.88	&	744.15	&	746.22	\\ [0.1 ex]
	&	${T}$	&	743.80	&	744.98	&	744.45	&	746.62	\\ [0.1 ex]
	&	Rank	&	1	&	3	&	2	&	4	\\ [0.1 ex]
    \midrule \multirow{4}{*}{Ionosphere} 	&	${T_1}$	&	0.17	&	0.14	&	0.31	&	0.39	\\ [0.1 ex]
	&	${T_2}$	&	692.70	&	684.53	&	683.68	&	684.99	\\ [0.1 ex]
	&	${T}$	&	692.87	&	684.68	&	683.98	&	685.38	\\ [0.1 ex]
	&	Rank	&	4	&	2	&	1	&	3	\\ [0.1 ex]
    \midrule \multirow{4}{*}{Lung Cancer}	&	${T_1}$	&	0.20	&	0.13	&	0.30	&	0.54	\\ [0.1 ex]
	&	${T_2}$	&	667.91	&	657.81	&	658.03	&	660.94	\\ [0.1 ex]
	&	${T}$	&	668.10	&	657.95	&	658.33	&	661.48	\\ [0.1 ex]
	&	Rank	&	4	&	1	&	2	&	3	\\ [0.1 ex]
    \midrule \multirow{4}{*}{Sonar}	&	${T_1}$	&	0.15	&	0.17	&	0.26	&	0.36	\\ [0.1 ex]
	&	${T_2}$	&	634.14	&	638.81	&	634.58	&	638.25	\\ [0.1 ex]
	&	${T}$	&	634.30	&	638.98	&	634.85	&	638.60	\\ [0.1 ex]
	&	Rank	&	1	&	4	&	2	&	3	\\ [0.1 ex]
    \midrule \multirow{4}{*}{Ozone}	&	${T_1}$	&	0.24	&	0.10	&	0.43	&	0.53	\\ [0.1 ex]
	&	${T_2}$	&	1622.54	&	1564.31	&	1327.62	&	1247.42	\\ [0.1 ex]
	&	${T}$	&	1622.78	&	1564.41	&	1328.06	&	1247.94	\\ [0.1 ex]
	&	Rank	&	4	&	3	&	2	&	1	\\ [0.1 ex]
    \midrule \multirow{4}{*}{Libras}	&	${T_1}$	&	0.19	&	0.17	&	0.40	&	0.64	\\ [0.1 ex]
	&	${T_2}$	&	764.47	&	784.85	&	775.89	&	748.75	\\ [0.1 ex]
	&	${T}$	&	764.66	&	785.02	&	776.29	&	749.39	\\ [0.1 ex]
	&	Rank	&	2	&	4	&	3	&	1	\\ [0.1 ex]
    \midrule \multirow{4}{*}{Hill}	&	${T_1}$	&	0.20	&	0.13	&	0.33	&	0.56	\\ [0.1 ex]
	&	${T_2}$	&	829.65	&	834.56	&	819.26	&	772.21	\\ [0.1 ex]
	&	${T}$	&	829.85	&	834.69	&	819.59	&	772.77	\\ [0.1 ex]
	&	Rank	&	3	&	4	&	2	&	1	\\ [0.1 ex]
    \midrule \multirow{4}{*}{Musk-1}	&	${T_1}$	&	0.27	&	0.14	&	0.38	&	0.78	\\ [0.1 ex]
	&	${T_2}$	&	889.12	&	881.33	&	914.55	&	824.34	\\ [0.1 ex]
	&	${T}$	&	889.39	&	881.48	&	914.93	&	825.12	\\ [0.1 ex]
	&	Rank	&	3	&	2	&	4	&	1	\\ [0.1 ex]
    \midrule \multirow{4}{*}{Arrhythmia}	&	${T_1}$	&	0.31	&	0.21	&	0.52	&	1.01	\\ [0.1 ex]
	&	${T_2}$	&	982.07	&	979.67	&	989.75	&	840.62	\\ [0.1 ex]
	&	${T}$	&	982.38	&	979.87	&	990.27	&	841.63	\\ [0.1 ex]
	&	Rank	&	3	&	2	&	4	&	1	\\ [0.1 ex]
    \midrule \multirow{4}{*}{LSVT}	&	${T_1}$	&	0.21	&	0.17	&	0.56	&	1.08	\\ [0.1 ex]
	&	${T_2}$	&	739.06	&	723.87	&	737.56	&	714.54	\\ [0.1 ex]
	&	${T}$	&	739.27	&	724.04	&	738.12	&	715.62	\\ [0.1 ex]
	&	Rank	&	4	&	2	&	3	&	1	\\ [0.1 ex]
    \midrule \multirow{4}{*}{SECOM}	&	${T_1}$	&	0.34	&	0.31	&	0.76	&	1.80	\\ [0.1 ex]
	&	${T_2}$	&	5879.65	&	5361.80	&	5982.40	&	5516.33	\\ [0.1 ex]
	&	${T}$	&	5879.99	&	5362.12	&	5983.16	&	5518.13	\\ [0.1 ex]
	&	Rank	&	3	&	1	&	4	&	2	\\ [0.1 ex]
    \midrule \multicolumn{2}{c}{\textbf{Average Rank}}			&	2.75	&	2.75	&	2.69	&	1.81	\\ [0.1 ex]
    \multicolumn{2}{c}{\textbf{Overall Rank}}			&	3	&	3	&	2	&	1	\\ [0.1 ex]
    
    \bottomrule
    \end{tabular}
    \end{threeparttable}
    }
    \end{minipage} 
\end{table}

\subsection{Dynamic Search Behavior}

\begin{figure}[t]
\centering
\begin{subfigure}{.24\textwidth}
  \centering
  \includegraphics[width=\textwidth]{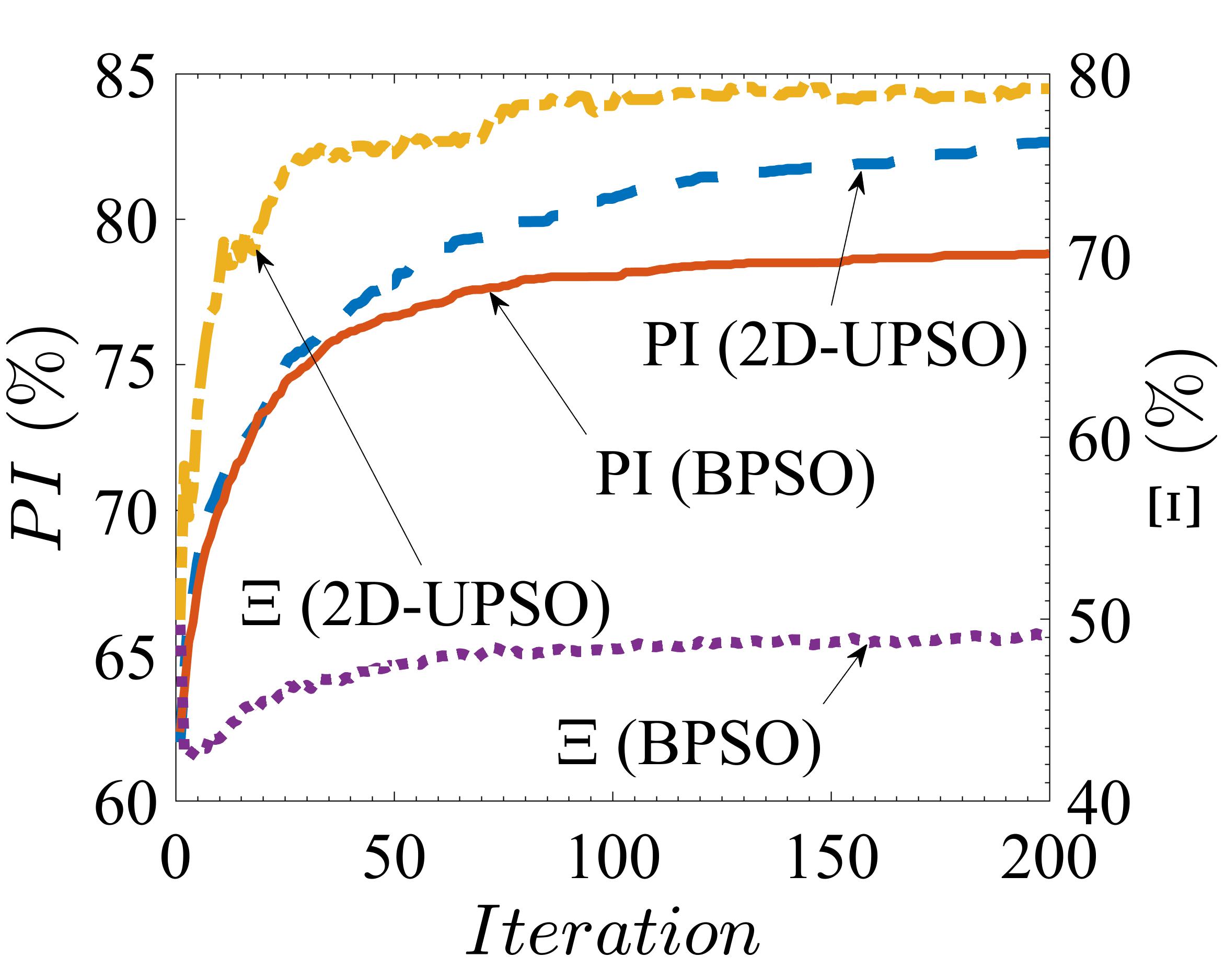}
  \caption{LSVT + NB}
  \label{f:LSVT_nb}
\end{subfigure}
\begin{subfigure}{.24\textwidth}
  \centering
  \includegraphics[width=\textwidth]{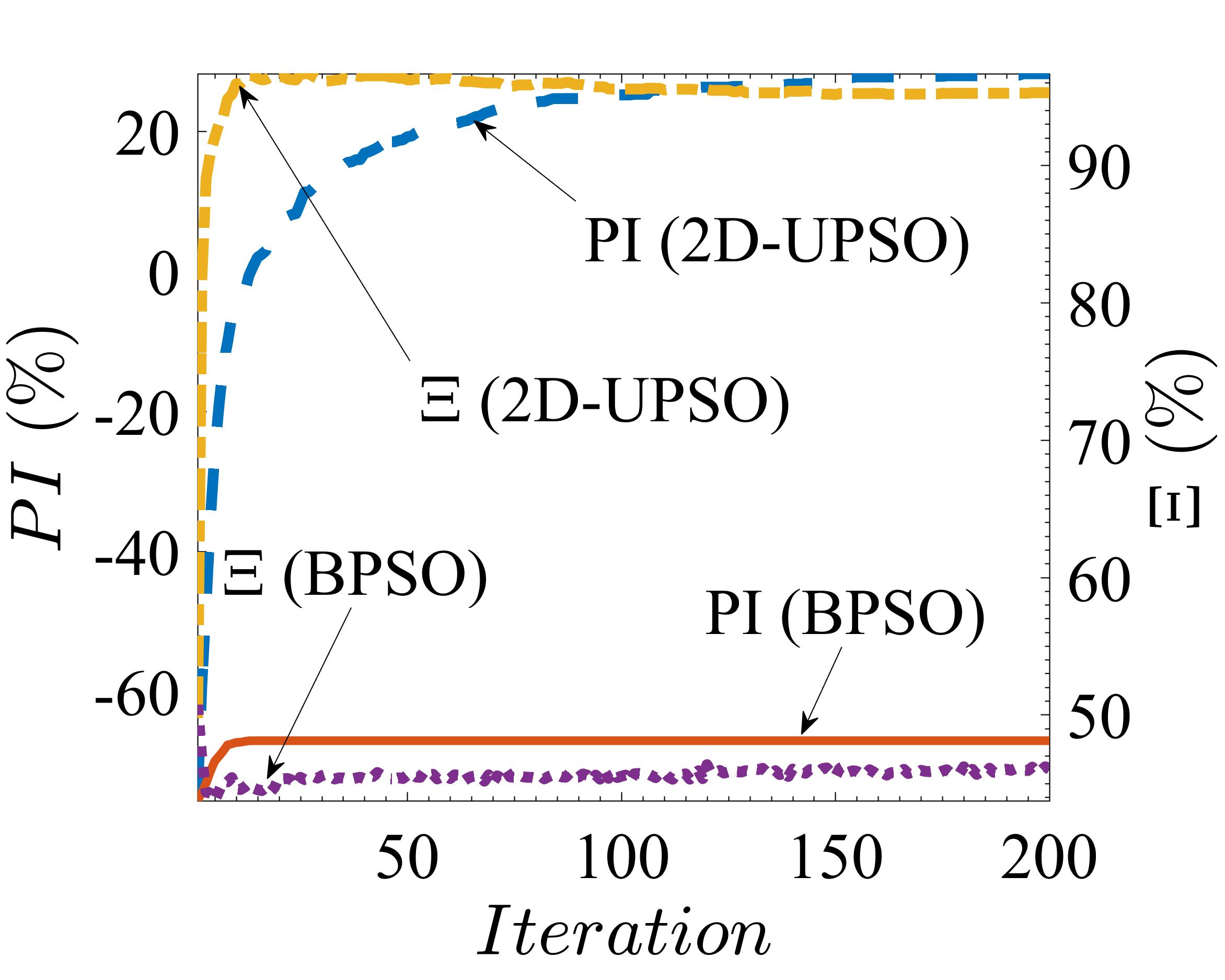}
  \caption{LSVT + k-NN}
  \label{f:LSVT_knn}
\end{subfigure}%
\begin{subfigure}{.24\textwidth}
  \centering
  \includegraphics[width=\textwidth]{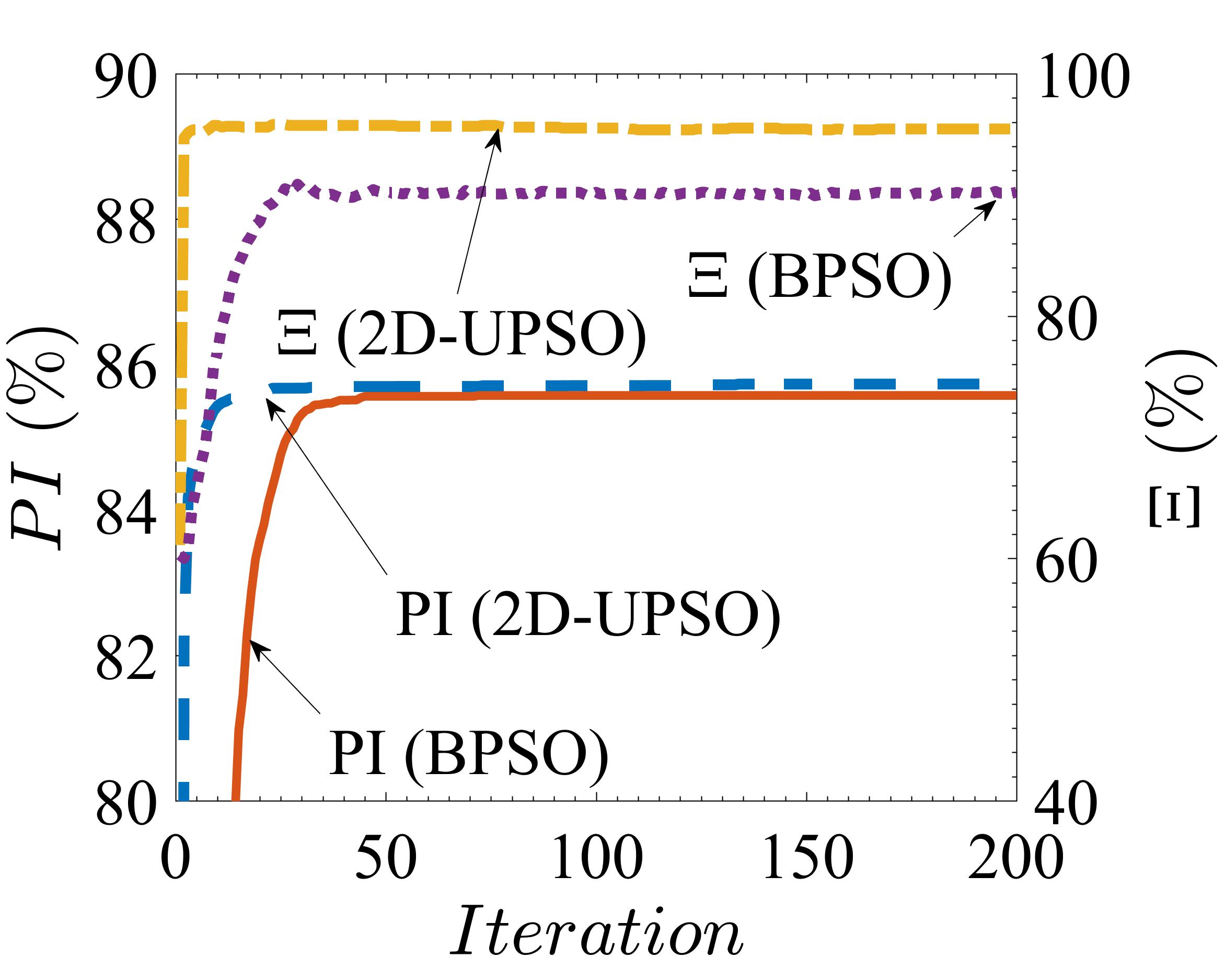}
  \caption{Ionosphere + NB}
  \label{f:iono_nb}
\end{subfigure}
\begin{subfigure}{.24\textwidth}
  \centering
  \includegraphics[width=\textwidth]{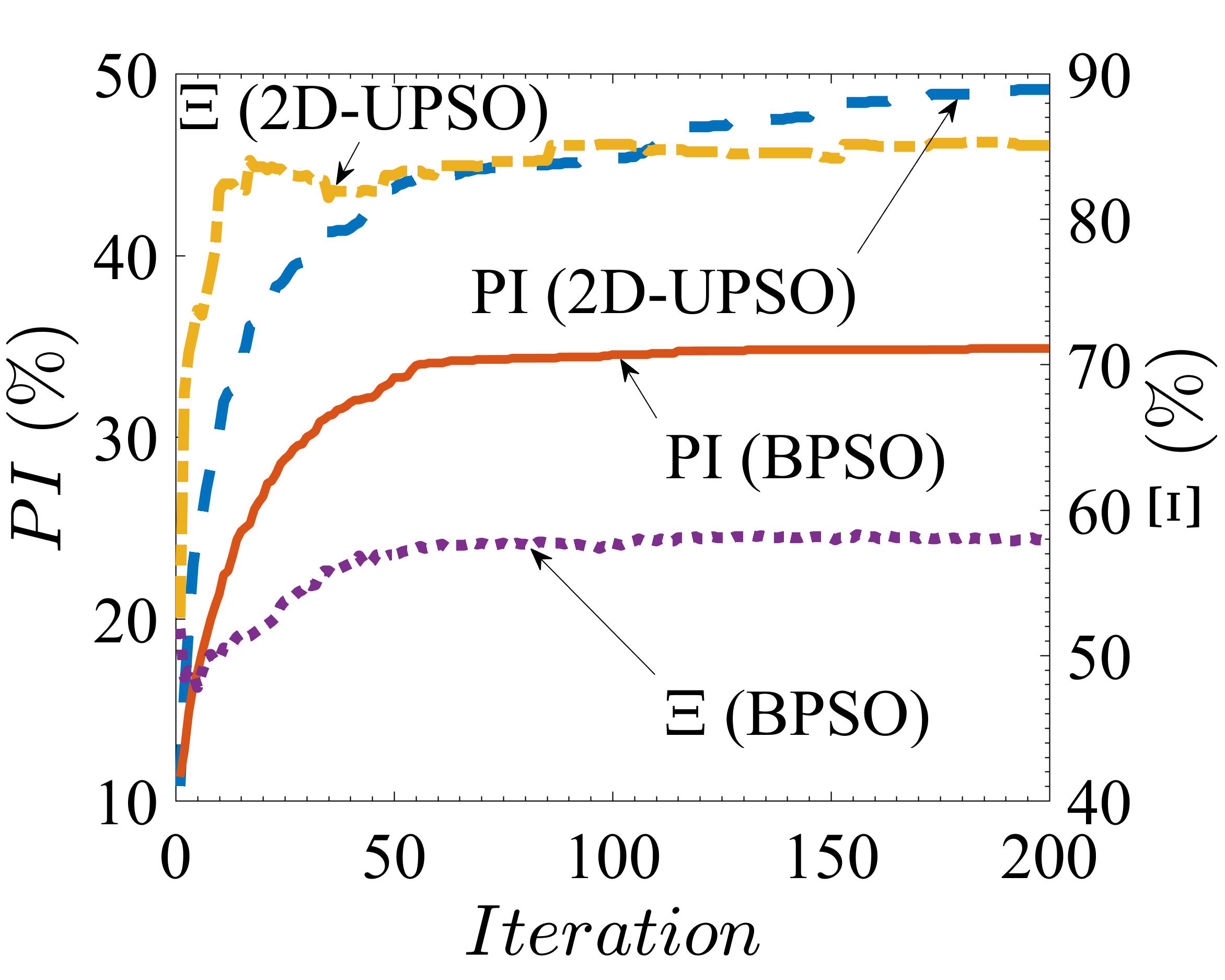}
  \caption{Ionosphere + k-NN}
  \label{f:iono_knn}
\end{subfigure}
\caption{Dynamic search behavior of 2D-UPSO and BPSO}
\label{f:pix}
\end{figure}

In order to have better insight into the search dynamics of the algorithms, both $PI (\%)$ and $\Xi(\%)$ of the best particle ($gbest$) were recorded at each iteration for few datasets. Note that since amongst the compared algorithms, BPSO provides the best overall performance, it would interesting to compare its performance with 2D-UPSO. The average variation in $PI(\%)$ and $\Xi(\%)$ obtained from 2D-UPSO and BPSO algorithms, after 40 runs are shown in Fig. \ref{f:pix} for two datasets, `LSVT' and `Ionosphere'.

As seen from Fig. \ref{f:LSVT_nb}-\ref{f:iono_knn}, throughout the search process, $\Xi(\%)$ obtained with 2D-UPSO is significantly higher than BPSO. This indicates that the cardinality ($\xi$) of the $gbest$ obtained from 2D-UPSO is significantly smaller. The superior performance of the 2D-UPSO can be ascribed to the exploitation of the cardinality information in the proposed 2D-learning approach. This distinctive quality of the 2D learning approach often discards the redundant features without the need of any additional heuristic information. This results smaller cardinality and significant reduction in the classification error. Further, it is to be noted that the improvement in $\Xi(\%)$ is not \textit{monotonic} as seen in Fig. \ref{f:pix}. Since the primary search objective is to obtain better classification performance, in each iteration the \textit{best particle} or $gbest$ is updated on the basis of its classification performance. Therefore, it does not necessarily have same or smaller cardinality compared to the previous iteration resulting non-monotonic behavior.

The similar improvements are observed in $PI(\%)$ with 2D-UPSO. For both the datasets, as seen in Fig. \ref{f:LSVT_nb}-\ref{f:iono_knn}, throughout the search the $PI(\%)$ obtained by 2D-UPSO is comparatively higher. Further, since LSVT is \textit{ill-defined} problem, the \textit{curse of dimensionality} is more likely to affect the classifier. This can be observed in Fig. \ref{f:LSVT_knn}, when k-NN is used as the induction algorithm. As discussed earlier, the results indicate that the search landscape defined by LSVT and k-NN classifier is quite deceptive. For this reason, during initial search stages (Fig. \ref{f:LSVT_knn}), both 2D-UPSO and BPSO failed to find a subset with classification performance better than $U$, this results in a negative value of $PI(\%)$ in the beginning. However, as the search progressed (fig. \ref{f:LSVT_knn}), 2D-UPSO quickly attained positive $PI(\%)$ whereas BPSO was unable to come out from local minima and failed to provide performance improvement over the original subset, $U$.

\section{Conclusions}
\label{sec:conclusion}

The performance of a new 2D learning framework has been investigated for feature selection problem. One of the important characteristic of this framework is that it exploits the information about the cardinality and integrates it into the search process by extending the dimension of the velocity. The efficacy of the proposed framework is demonstrated by adapting one of the popular PSO variant, UPSO, by considering two different induction algorithms,\textit{i.e.}, NB and k-NN. The time-complexity of this algorithm has been investigated and is compared with many existing algorithms. The comparative study performed on wide variety of datasets demonstrate that, for most of the datasets, the 2D-UPSO could find \textit{smaller} feature subset with \textit{better} classification performance in \textit{shorter} running time and provides overall best performance irrespective of the induction algorithm being used. 

However, the current study is focused on the \textit{single objective} approach to the feature selection where the main focus is to improve the classification performance. In some cases, a designer may be interested to trade-off the classification performance in favor of smaller subset. In such scenario, a multi-objective search criteria, with focus on both \textit{classification accuracy} and \textit{cardinality}, can be very helpful. Further, in this study the search parameters such as inertia weight ($\omega$), acceleration constant ($c_1,c_2$), unification factor ($u$) were selected empirically. A detailed study on parameter control, similar to the existing studies in the continuous domain, can further improve the search performance. This could be the subject of further research.

\bibliographystyle{elsarticle-num}


\end{document}